\documentclass[journal, 10pt, twocolumn]{IEEEtran}
\usepackage{cite}
\ifCLASSINFOpdf
\else
   \usepackage[dvips]{graphicx}
   \DeclareGraphicsExtensions{.eps}
\fi
\usepackage[cmex10]{amsmath}
\usepackage{algorithmic}
\usepackage{amssymb}

\usepackage{array}
\usepackage{mdwmath}
\usepackage{mdwtab}
\usepackage{eqparbox}
\usepackage{multirow}

\usepackage{fixltx2e}

\usepackage{enumerate}

\usepackage{url}
\usepackage{tabularx}

\begin{document}
%
% paper title
% can use linebreaks \\ within to get better formatting as desired
%\title{Learning Doubly Sparse Transforms \\ for Image Processing}
\title{Efficient Sum of Outer Products Dictionary Learning (SOUP-DIL) -- The $\ell_0$ Method}

%Based Magnetic Resonance Image Reconstruction

%Sparsifying Transform Learning with Closed-Form Optimal Updates and Convergence Guarantees

%
%
% author names and IEEE memberships
% note positions of commas and nonbreaking spaces ( ~ ) LaTeX will not break
% a structure at a ~ so this keeps an author's name from being broken across
% two lines.
% use \thanks{} to gain access to the first footnote area
% a separate \thanks must be used for each paragraph as LaTeX2e's \thanks
% was not built to handle multiple paragraphs
%

\author{Saiprasad~Ravishankar,~\IEEEmembership{Member,~IEEE,}~Raj~Rao~Nadakuditi,~\IEEEmembership{Member,~IEEE,}\\ and~Jeffrey~A.~Fessler,~\IEEEmembership{Fellow,~IEEE}% <-this % stops a space
%\thanks{Copyright (c) 2014 IEEE. Personal use of this material is permitted. However, permission to use this material for any other purposes must be obtained from the IEEE by sending a request to pubs-permissions@ieee.org.}
%\thanks{This work was first submitted to the IEEE Transactions on Signal Processing on April 02, 2014. It was accepted in the current form on January 09, 2015.}
\thanks{This work was supported in part by the following grants: ONR grant N00014-15-1-2141, DARPA Young Faculty Award  D14AP00086, ARO MURI grants  W911NF-11-1-0391 and 2015-05174-05, NIH grant U01 EB018753, and a UM-SJTU seed grant.}
\thanks{S. Ravishankar, R. Nadakuditi,  and J. Fessler are with the Department of Electrical Engineering and Computer Science, University of Michigan, Ann Arbor, MI, 48109 USA emails: (ravisha, rajnrao, fessler)@umich.edu.}}
\maketitle

%The learnt dictionary is also constrained to be incoherent which typically improves the accuracy of  learning.
\begin{abstract}
The sparsity of natural signals and images in a transform domain or dictionary has been extensively exploited in several applications such as compression, denoising and inverse problems. More recently, data-driven adaptation of synthesis dictionaries has shown promise in many applications compared to fixed or analytical dictionary models. However, dictionary learning problems are typically non-convex and NP-hard, and the usual alternating minimization approaches for these problems are often computationally expensive, with the computations dominated by the NP-hard synthesis sparse coding step.
In this work, we investigate an efficient method for $\ell_{0}$ ``norm"-based dictionary learning by first approximating the training data set with a sum of sparse rank-one matrices and then using a block coordinate descent approach to estimate the unknowns. The proposed block coordinate descent algorithm involves efficient closed-form solutions. In particular, the sparse coding step involves a simple form of thresholding. We provide a convergence analysis for the proposed block coordinate descent approach.  Our numerical experiments show the promising performance and significant speed-ups provided by our method over the classical K-SVD scheme in sparse signal representation and image denoising.
%\boldmath
\end{abstract}

%In particular, the sparse coding step is usually solved using approximate greedy methods. In this work, we take a 

%We provide local convergence guarantees for our alternating algorithms. 

%Our adaptive well-conditioned transforms are shown to perform better in applications compared to adapted orthonormal ones.

%These transforms are a product of an adaptive sparse matrix and a fixed, fast analytic transform such as the DCT. 

% in MRI
% IEEEtran.cls defaults to using nonbold math in the Abstract.
% This preserves the distinction between vectors and scalars. However,
% if the journal you are submitting to favors bold math in the abstract,
% then you can use LaTeX's standard command \boldmath at the very start
% of the abstract to achieve this. Many IEEE journals frown on math
% in the abstract anyway.

% Note that keywords are not normally used for peerreview papers.


%Sparsifying transforms, Efficient solutions, Convergence guarantees, Image representation, Sparse representation,  Denoising,  Dictionary learning, Non-convex.

%\begin{IEEEkeywords}
%Sparsifying transforms, Compressed Sensing, Image representation, Denoising, Sparse representation, Dictionary learning, Magnetic image reconstruction.
%\end{IEEEkeywords}

% For peer review papers, you can put extra information on the cover
% page as needed:
% \ifCLASSOPTIONpeerreview
% \begin{center} \bfseries EDICS Category: 3-BBND \end{center}
% \fi
%
% For peerreview papers, this IEEEtran command inserts a page break and
% creates the second title. It will be ignored for other modes.

\IEEEpeerreviewmaketitle

\section{Introduction} \label{sec1}

The sparsity of natural signals and images in a transform domain or dictionary has been extensively exploited in several applications such as compression \cite{jpg2}, denoising, compressed sensing and other inverse problems. Well-known models for sparsity include the synthesis, analysis  \cite{elmiru}, and the transform   \cite{tfcode, sabres} (or generalized analysis) models. More recently, the data-driven adaptation of sparse signal models has benefited many applications \cite{elad2, elad3, elad4, Mai, bresai, akd, sabres, doubsp2l, saiwen} compared to fixed or analytical models.
In this work, we focus on the data-driven adaptation of the synthesis model and present a simple and highly efficient algorithm with convergence analysis and applications. 
% In this two part work, we focus on the data-driven adaptation of the synthesis model and present highly efficient algorithms with convergence analysis and applications. In Part I, we focus entirely on a particular highly non-convex formulation for dictionary learning. Part II is devoted to the investigation of some multi-convex variants of the proposed problem. 
In the following, we first briefly review the topic of synthesis dictionary learning before summarizing the contributions of this work.

\subsection{Synthesis Model and Dictionary Learning}\label{sec1a} 

The well-known synthesis model suggests that a signal $ y  \in \mathbb{R}^{n} $ is approximately a linear combination of a small subset of atoms or columns of a dictionary $D  \in \mathbb{R}^{n \times J} $, i.e., $y = D x + e$ with $x \in \mathbb{R}^{J} $ sparse, and $e$ is assumed to be a small modeling error or approximation error in the signal domain \cite{ambruck}. We say that $x \in \mathbb{R}^{J} $ is sparse if $\left \| x \right \|_{0}\ll n$, where the $\ell_0$ ``norm" counts the number of non-zero entries in $x$. Since different candidate signals may be approximately spanned by different subsets of columns in the dictionary $D$, the synthesis model is also known as a union of subspaces model \cite{vidal2, elhamm}. When $n=J$ and $D$ is full rank, it is a basis. Else when $J>n$, $D$ is called an overcomplete dictionary. Because of their richness, overcomplete dictionaries can provide highly sparse (i.e., with few non-zeros) representations of data and are popular.

For a given signal $y$ and dictionary $D$, the process of finding a sparse representation $x$ involves solving the well-known synthesis sparse coding problem. This problem is to  minimize $\left \| y-Dx \right \|_{2}^{2}$ subject to  $\left \| x \right \|_{0}\leq s$, where $s$ is some set sparsity level. The synthesis sparse coding problem is NP-hard (Non-deterministic Polynomial-time hard) \cite{npb, npa}. Numerous algorithms \cite{pati, mp2, RaoFocus, Harikumar:Bresler-96, chen2,  befro, Needell2, wei} including greedy and relaxation algorithms have been proposed for this problem. While some of these algorithms are guaranteed to provide the correct solution under certain conditions, these conditions are often restrictive and violated in applications. Moreover, these sparse coding algorithms typically tend to be computationally expensive for large-scale problems.

More recently, the data-driven adaptation of synthesis dictionaries, called dictionary learning, has been investigated in several works \cite{ols, eng, elad, Yagh, Mai}. 
%has been shown to be advantageous in applications compared to fixed or analytical dictionaries such as wavelets. Several applications have been explored for dictionary learning including 
Dictionary learning has  shown promise in several applications including
compression, denoising, inpainting, deblurring, demosaicing, super-resolution, and classification \cite{elad2, Bryt20, elad3, elad5, elad6, mai23, super2, irami, Kongwang, ladav233}. It has also been demonstrated to be useful in inverse problems such as those in tomography \cite{gusup}, and magnetic resonance imaging (MRI) \cite{bresai, symul, wangying, huangbays}.

%Synthesis dictionary learning has been investigated in several works \cite{ols, eng, elad, Yagh, Mai}. 

Given a collection of training signals $\left \{ y_{i} \right \}_{i=1}^{N}$ that are represented as columns of the matrix $ Y \in \mathbb{R}^{n \times N} $, the dictionary learning problem is often formulated as follows \cite{elad}
\begin{align} 
\nonumber (\mathrm{P0})\; & \min_{D,X}\: \left \| Y-DX \right \|_{F}^{2}\; \: \mathrm{s.t.}\; \:  \left \| x _{i} \right \|_{0}\leq s\; \: \forall \,  i, \, \left \| d_{j} \right \|_2 =1 \,\forall \, j.
\end{align}
Here, $d_{j}$ and $x_{i}$ denote the columns of the dictionary $D \in \mathbb{R}^{n \times J}$ and sparse code matrix $X \in \mathbb{R}^{J \times N}$, respectively, and $s$ denotes the maximum sparsity level (non-zeros in representations $x_i$) allowed for each training signal. The columns of the dictionary are constrained to have unit norm to avoid the scaling ambiguity \cite{kar}. Variants of Problem (P0) include replacing the $\ell_0$ ``norm" for sparsity with an $\ell_1$ norm or an alternative sparsity criterion, or enforcing additional properties (e.g., incoherence \cite{barchi1, irami}) for the dictionary $D$, or solving an online version (where the dictionary is updated sequentially as new training signals arrive) of the problem \cite{Mai}.

Algorithms for Problem (P0) or its variants \cite{eng, elad, Yagh, zibul, skret, Mai, ophel4, kmeansgen, smith1, sadeg2, segh2, behes1, bao1} typically alternate in some form between a \emph{sparse coding step} (updating $X$), and a \emph{dictionary update step} (solving for $D$). Some of these algorithms (e.g., \cite{elad, smith1, segh2}) also partially update $X$ in the dictionary update step. A few recent methods attempt to solve for $D$ and $X$ jointly in an iterative fashion \cite{directdl11, hawesep22}.
The K-SVD method \cite{elad} has been particularly popular in numerous applications \cite{elad2, elad3, elad4,  Bryt20, bresai, symul}.
However, Problem (P0) is highly non-convex and NP-hard, and most dictionary learning approaches lack proven convergence guarantees. Moreover, the algorithms for (P0) tend to be computationally expensive (particularly alternating-type algorithms), with the computations usually dominated by the synthesis sparse coding step.

Some recent works  \cite{spel2b, agra1, arora1, yint3, bao1, agra2} have studied the convergence of (specific) synthesis dictionary learning algorithms.
However, these dictionary learning methods have not been demonstrated to be useful in applications such as image denoising. Bao et al. \cite{bao1} find that their method, although a fast proximal scheme, denoises less effectively than the K-SVD method \cite{elad2}. Many prior works use restrictive assumptions (e.g., noiseless data, etc.) for their convergence results.

%(typically $0.1-0.3$ dB worse)

\subsection{Contributions} \label{sec1b} 

This work focuses on synthesis dictionary learning and investigates a dictionary learning problem with an $\ell_{0}$ sparsity penalty instead of constraints. %In particular, an $\ell_0$ penalty for sparsity is explored in this work. 
%In Part II, we analyze the usefulness of some multi-convex (i.e., convex with respect to blocks of variables) variants of the problem proposed in this work. 
Our approach first models the training data set as an approximate sum of sparse rank-one matrices (or outer products). Then, we use a simple and exact block coordinate descent approach to estimate the factors of the various rank-one matrices. In particular, the sparse coding step in the algorithm uses a simple form of thresholding, and the dictionary atom update step involves a matrix-vector product that is computed efficiently. We provide a convergence analysis of the proposed block coordinate descent method.
%The convergence results are further strengthened for the methods discussed in Part II. 
Our numerical experiments demonstrate the promising performance and significant speed-ups provided by our method over the classicial K-SVD dictionary learning scheme in sparse signal representation and image denoising. Importantly, our algorithm is similar to just the dictionary update step of K-SVD, and yet performs comparably or much better and faster than K-SVD in applications.

%At the time of writing this paper, we became aware of a very recent similar dictionary learning approach \cite{sadeg33}. 

Our work shares similarities with a recent dictionary learning approach \cite{sadeg33} that exploits a sum of outer products model for the training data. However, the specific problem formulation and algorithm studied in this work differ from the prior work \cite{sadeg33}. 
%The approach in \cite{sadeg33} is similar to one of the schemes (that with an $\ell_1$ penalty for sparsity) in Part II. However, unlike the related method in Part II, the prior method \cite{sadeg33} does not involve multi-convexity. 
Importantly, unlike the previous approach \cite{sadeg33}, we provide a detailed convergence analysis of the proposed algorithm and also demonstrate its usefulness (both in terms of speed and quality of results) in sparse representation of natural signals and in image denoising. The sum of outer products model and related algorithm can accomodate various structures or constraints on the dictionary or sparse codes.

% However, the dictionary learning problems are typically non-convex and NP-hard, and the alternating minimization approaches usually adoped to solve these problems tend to be computationally expensive, with the computations dominated by the NP-hard synthesis sparse coding step.

\subsection{Organization} \label{sec1c} 

The rest of this paper is organized as follows. Section \ref{sec2} discusses the problem formulation for dictionary learning, along with some potential alternatives. Section \ref{sec3} presents our dictionary learning algorithm and discusses its computational properties. Section \ref{sec4} presents a convergence analysis for the proposed algorithm. Section \ref{sec5} illustrates the empirical convergence behavior of our method and demonstrates its usefulness for sparse signal representation and image denoising. Section \ref{sec6} presents conclusions and proposals for future work.

\section{Problem Formulations}
\label{sec2}

\subsection{$\ell_{0}$ Penalized Formulation} \label{sec2a} 

We consider a sparsity penalized variant of Problem (P0) \cite{bao1}. Specifically, by replacing the sparsity constraints in (P0) with an $\ell_0$ penalty $\sum_{i=1}^{N} \left \| x_{i} \right \|_{0}$ and introducing a variable $C = X^{T} \in \mathbb{R}^{N \times J}$, where $(\cdot)^{T}$ denotes the matrix transpose operation, we arrive at the following formulation:
\begin{align}  \label{eqop4}
& \min_{D,C}\: \left \| Y-DC^{T} \right \|_{F}^{2} + \lambda^{2} \left \| C \right \|_{0} \; \: \mathrm{s.t.}\; \: \left \| d_{j} \right \|_2 =1 \,\forall \, j.
\end{align}
where $\left \| C \right \|_{0}$ counts the number of non-zeros in the matrix $C$, and $\lambda^{2} >0$ is a weight for the sparsity penalty that controls the number of non-zeros in the sparse representation.

Next, we express the matrix $DC^{T}$ in \eqref{eqop4} as a sum of (sparse) rank-one matrices or outer products $\sum_{j=1}^{J} d_{j}c_{j}^{T}$, where $c_{j}$ is the $j$th column of $C$. This is a natural representation of the training data $Y$ because it separates out the contributions of the various atoms in representing the data. It may also provide a natural way to set the number of atoms (degrees of freedom) in the dictionary. In particular, atoms of a dictionary whose contributions to the data ($Y$) representation error or modeling error are small could be dropped. Such a Sum of OUter Products (SOUP) representation has been exploited in previous dictionary learning algorithms  \cite{elad, smith1}. With this model, we rewrite \eqref{eqop4} as follows, where $\left \| C \right \|_{0} = \sum_{j=1}^{J} \left \| c_{j} \right \|_{0}$:
\begin{align} 
\nonumber (\mathrm{P1})\; & \min_{\left \{ d_{j},c_{j} \right \}}\: \begin{Vmatrix}
Y- \sum_{j=1}^{J} d_{j}c_{j}^{T}
\end{Vmatrix}_{F}^{2} + \lambda^{2} \sum_{j=1}^{J} \left \| c_{j} \right \|_{0} \\
\nonumber & \; \; \; \mathrm{s.t.}\; \: \left \| d_{j} \right \|_2 =1, \, \left \| c_{j} \right \|_{\infty} \leq L \,\forall \, j.
\end{align}
As in Problem (P0), the matrix $d_{j}c_{j}^{T}$ in (P1) is invariant to joint scaling of $d_{j}$ and $c_{j}$ as $\alpha d_{j}$ and $(1/\alpha) c_{j}$, for $\alpha \neq 0$. The constraint $\left \| d_{j} \right \|_2 =1$ helps remove this scaling ambiguity. 
%In addition, we have enforced the constraint $\left \| c_{k} \right \|_{\infty} \leq L$, with $L>0$ \cite{bao1}, in (P1).  When the optimal dictionary in (P1) has a column $d_{j}$ that repeats, then the outer product expansion of $Y$ in (P1) can potentially have both the terms $d_{j}c_{j}^{T}$ and $-d_{j}c_{j}^{T}$ with $c_{j}$ that is highly sparse, and the objective would be invaraint to scaling of $c_{j}$. The $\ell_{\infty}$ constraint on $c_{k}$ prevents such a scenario.  
We also enforce the constraint $\left \| c_{j} \right \|_{\infty} \leq L$, with $L>0$, in (P1) \cite{bao1} (e.g., $L=\left \| Y \right \|_{F}$).
Consider a dictionary $D$ that has a column $d_{j}$ that repeats. Then, in this case, the outer product expansion of $Y$ in (P1) could have both the terms $d_{j}c_{j}^{T}$ and $-d_{j}c_{j}^{T}$ with $c_{j}$ that is highly sparse, and the objective would be invariant to (arbitrarily) large scalings\footnote{Such degenerate representations of $Y$, however, cannot be minimizers in the problem because they simply increase the $\ell_{0}$ sparsity penalty without affecting the fitting error (the first term) in the cost.} of $c_{j}$ (i.e., non-coercive objective).
The $\ell_{\infty}$ constraints on the columns of $C$ (that constrain the magnitudes of entries of $C$) prevent such a scenario.
%eliminate such representations from the set of feasible representations in (P1).
Problem (P1) is designed to learn the factors $\left \{ d_{j} \right \}_{j=1}^{J}$ and $\left \{ c_{j} \right \}_{j=1}^{J}$ that enable the best SOUP sparse representation of $Y$.

% although feasible in the absence of the $\ell_{\infty}$ constraints in (P1)

%We do not constrain the magnitudes of the entries in $C$ \cite{bao1}, as we did not find any advantages in doing so.

\subsection{Alternative Formulations} \label{sec2b} 

Several variants of the SOUP learning Problem (P1) could be constructed, with interesting effects.  For example, the $\ell_0$ ``norm" for sparsity could be replaced by the $\ell_1$ norm \cite{sadeg33}. %that turn out to be multi-convex problems and lead to stronger convergence results for the related algorithms.
Another interesting alternative to (P1) involves enforcing $p$-block-orthogonality constraints on the dictionary $D$. The dictionary in this case is split into blocks (instead of individual atoms), each of which has $p$ atoms that are orthogonal to each other. For $p=2$, the constraints take the form  $d_{2j-1}^{T}d_{2j} = 0$, $1\leq j \leq J/2$. In the extreme (more constrained) case of $p=n$, the dictionary would be made of several (square) orthonormal blocks.
For tensor-type data, Problem (P1) can be modified by enforcing the dictionary atoms to be in the form of a Kronecker product.
The algorithm proposed in Section \ref{sec3} can be easily extended to accomodate such variants of Problem (P1).
We do not fully explore such alternatives in this work due to space constraints, and a more detailed investigation of these will be presented elsewhere.

\newtheorem{theorem}{Theorem}
\newtheorem{cor}{Corollary}
\newtheorem{lem}{Lemma}
\newtheorem{prop}{Proposition}

\section{Algorithm and Properties}
\label{sec3}  

\subsection{Algorithm} \label{sec3a} 

We propose a block coordinate descent method to estimate the unknown variables in Problem (P1). For each $j$ ($1 \leq j \leq J$), our algorithm has two key steps.
First, we solve (P1) with respect to $c_{j}$ keeping all the other variables fixed. We refer to this step as the \emph{sparse coding step} in our method. Once $c_{j}$ is updated, we solve (P1) with respect to $d_{j}$ keeping all other variables fixed. This step is referred to as the \emph{dictionary atom update step} or simply \emph{dictionary update step}. The algorithm updates the factors of the various rank-one matrices one-by-one. We next describe the sparse coding and dictionary atom update steps.

\subsubsection{Sparse Coding Step} \label{sec3a1}
Minimizing (P1) with respect to $c_{j}$ leads to the following non-convex problem, where $E_{j} \triangleq Y - \sum_{k\neq j} d_{k}c_{k}^{T}$ is a fixed matrix based on the most recent values of all other atoms and coefficients:
\begin{equation} \label{eqop5}
\min_{c_{j}} \; \begin{Vmatrix}
E_{j} - d_{j}c_{j}^{T}
\end{Vmatrix}_{F}^{2} + \lambda^{2} \left \| c_{j} \right \|_{0}  \;\; \mathrm{s.t.}\; \: \left \| c_{j} \right \|_{\infty} \leq L.
\end{equation}
The following proposition provides the solution to Problem \eqref{eqop5}, where the hard-thresholding operator $H_{\lambda} (\cdot)$ is defined as
\begin{equation} \label{equ88ch4}
\left ( H_{\lambda} (b) \right )_{i}=\left\{\begin{matrix}
 0,& \;\;\left | b_{i} \right | < \lambda \\
b_{i},  & \;\;\left | b_{i} \right | \geq \lambda 
\end{matrix}\right.
\end{equation}
with $b \in \mathbb{R}^{N}$, and the subscript $i$ indexes vector entries. We assume that the bound $L > \lambda$ and let $1_{N}$ denote a vector of ones of length $N$. The operation ``$\odot$" denotes element-wise multiplication, $\mathrm{sign}(\cdot)$ computes the signs of the elements of a vector, and $z= \min(a, u)$ for vectors $a, u \in \mathbb{R}^{N}$ denotes the element-wise minimum operation, i.e., $z_{i}= \min(a_{i}, b_{i})$, $1 \leq i \leq N$.  

\begin{prop}\label{prop1} \vspace{0.02in}
Given $E_{j} \in \mathbb{R}^{n \times N}$ and $d_{j} \in \mathbb{R}^{n}$, and assuming $L > \lambda$,  a global minimizer of the sparse coding problem \eqref{eqop5} is obtained by the following truncated hard-thresholding operation:
\begin{equation} \label{tru1ch4}
%\hat{c}_{j} =  H_{\lambda} \left ( E_{j}^{T}d_{j} \right )
\hat{c}_{j} =  \min\left ( \begin{vmatrix}
H_{\lambda} \left ( E_{j}^{T}d_{j} \right )
\end{vmatrix}, L 1_{N} \right ) \, \odot \, \mathrm{sign}\left ( H_{\lambda} \left ( E_{j}^{T}d_{j} \right ) \right )
\end{equation}
The minimizer of \eqref{eqop5} is unique if and only if the vector $E_{j}^{T}d_{j}$ has no entry with a magnitude of $\lambda$. 
\end{prop}

\hspace{0.1in} \textit{Proof:} 
First, for a vector $d_{j}$ that has unit $\ell_2$ norm, we have the following equality
\begin{align} 
\nonumber & \begin{Vmatrix}
E_{j} - d_{j}c_{j}^{T}
\end{Vmatrix}_{F}^{2}= \left \| E_{j} \right \|_{F}^{2} +  \left \| c_{j} \right \|_{2}^{2} - 2 c_{j}^{T} E_{j}^{T} d_{j} \\
& \;\;\;\;= \left \| c_{j} - E_{j}^{T} d_{j} \right \|_{2}^{2} + \left \| E_{j} \right \|_{F}^{2} - \left \| E_{j}^{T}d_{j} \right \|_{F}^{2}
\label{eqop8}
\end{align}
By substituting \eqref{eqop8} into \eqref{eqop5},
%replacing the term $\begin{Vmatrix} E_{j} - d_{j}c_{j}^{T} \end{Vmatrix}_{F}^{2}$ in the objective of Problem \eqref{eqop5} with the right hand side above, 
it is clear that  Problem \eqref{eqop5} is equivalent to %finding the minimizer in the following problem:
\begin{equation} \label{eqop5b}
\min_{c_{j}} \; \left \| c_{j} - E_{j}^{T} d_{j} \right \|_{2}^{2}  + \lambda^{2} \left \| c_{j} \right \|_{0} \;\; \mathrm{s.t.}\; \: \left \| c_{j} \right \|_{\infty} \leq L.
\end{equation}
%This problem has a simple solution obtained by hard thresholding (cf. \cite{sbclsTS2}, for example). For completeness, we briefly derive this result here.
Define $b \triangleq  E_{j}^{T} d_{j}$. %and let subscript $i$ index the entries of a vector. 
Then, the objective in \eqref{eqop5b} simplifies to $\sum_{i=1}^{N} \left \{ \left | c_{ji} - b_{i} \right |^{2}  + \lambda^{2} \, \theta(c_{ji})
 \right \}$  with
\begin{equation} \label{bbt5apeq3}
 \theta \left ( a \right )=\left\{\begin{matrix}
 0,& \; \mathrm{if} \;\, a = 0 \\
1,  & \; \mathrm{if} \;\, a \neq 0
\end{matrix}\right.
\end{equation}

%Thus, each entry $c_{ji}$ of $c_{j}$ is obtained as %the minimizer in the following problem:
Therefore, we solve for each entry $c_{ji}$ of $c_{j}$ as
\begin{equation} \label{eqop10}
\hat{c}_{ji} = \underset{c_{ji}}{\arg \min}  \left | c_{ji} - b_{i} \right |^{2}  + \lambda^{2} \, \theta(c_{ji})   \;\; \mathrm{s.t.}\; \: \left | c_{ji} \right | \leq L.
\end{equation}

%Clearly, the optimal $\hat{c}_{ji}= 0$ whenever $b_{i}=0$. In general, if the optimal $\hat{c}_{ji}= 0$, then the optimal objective value in \eqref{eqop10} is   $b_{i}^{2}$. When the optimal $\hat{c}_{ji}\neq 0$, we consider two cases: a) when $\left | b_{i} \right | \leq L$, and b) when $\left | b_{i} \right | > L$.
%In case (a), the optimal objective value is $\lambda^{2}$ that is clearly obtained by setting $\hat{c}_{ji} = b_{i}$ in \eqref{eqop10}.
%In case (b), the optimal objective value is $\left ( L\, \mathrm{sign}(b_{i}) - b_{i} \right )^{2} + \lambda^{2}$, obtained by setting $\hat{c}_{ji} = L\, \mathrm{sign}(b_{i})$.
%Comparing the optimal objective values in cases (a) and (b) for when  $\hat{c}_{ji}\neq 0$ to the objective when $\hat{c}_{ji}= 0$, we have the following results.

It is straightforward to show that when $\left | b_{i} \right | \leq L$ (case (a)),
\begin{equation} \label{bbt5apeq5}
\hat{c}_{ji} =\left\{\begin{matrix}
 0,& \; \mathrm{if} \;\, b_{i}^{2} < \lambda^{2} \\
 b_{i},  & \; \mathrm{if} \;\, b_{i}^{2} > \lambda^{2}
\end{matrix}\right.
\end{equation}
%in \eqref{eqop10}
When $\left | b_{i} \right | = \lambda$ ($\lambda < L$), the optimal $\hat{c}_{ji}$ can be either $b_{i}$ or $0$ (non-unique), and both these settings achieve the minimum objective value $\lambda^{2}$ in \eqref{eqop10}.
Next, when $\left | b_{i} \right | > L$ (case (b)), we have
\begin{equation} \label{bbt5apeq5yu}
 \hat{c}_{ji} =\left\{\begin{matrix}
 0, & \mathrm{if} \;\, b_{i}^{2} < \left ( L\, \mathrm{sign}(b_{i}) - b_{i} \right )^{2} + \lambda^{2}\\
  L\, \mathrm{sign}(b_{i}),   & \; \mathrm{if} \;\, b_{i}^{2} > \left ( L\, \mathrm{sign}(b_{i}) - b_{i} \right )^{2} + \lambda^{2}
\end{matrix}\right.
\end{equation}
Since $L > \lambda$, clearly $b_{i}^{2} > \left ( L\, \mathrm{sign}(b_{i}) - b_{i} \right )^{2} + \lambda^{2}$ in \eqref{bbt5apeq5yu}. %Therefore, in this case $ \hat{c}_{ji}  =  L\, \mathrm{sign}(b_{i})$ is the only solution. 
%The results in the two cases above can be combined together and the optimal 

Thus, an optimal $\hat{c}_{ji}$ in \eqref{eqop10} is compactly written as $\hat{c}_{ji} =  \min\left ( \begin{vmatrix}
H_{\lambda} \left ( b_{i} \right )
\end{vmatrix}, L \right ) \, \cdot \, \mathrm{sign}\left ( b_{i} \right )$, thereby establishing \eqref{tru1ch4}. The condition for uniqueness of the sparse coding solution follows from the arguments for case (a) above.
%The above arguments establish that a particular solution to Problem \eqref{eqop5} can be obtained as $\hat{c}_{j} =  H_{\lambda} \left ( E_{j}^{T}d_{j} \right )$, and this solution is unique if and only $E_{j}^{T}d_{j}$ has no entry with magnitude $\lambda$.
%%%%%%%%%%%%
% Else, if $\hat{c}_{ji}\neq 0$, the optimal objective value is $\lambda^{2}$ obtained by setting $\hat{c}_{ji} = b_{i}$. 
%Comparing the objective values in these two cases, we have that 
$\;\;\; \blacksquare$

\subsubsection{Dictionary Atom Update Step} \label{sec3a2}

Minimizing (P1) with respect to $d_{j}$ leads to the following non-convex problem:
\begin{equation} \label{eqop6}
 \min_{d_{j}} \; \begin{Vmatrix}
E_{j} - d_{j}c_{j}^{T}
\end{Vmatrix}_{F}^{2}  \;\:\; \mathrm{s.t.}\; \: \left \| d_{j} \right \|_2 =1
\end{equation}
Proposition \ref{prop2} provides the closed-form solution for \eqref{eqop6}.

\begin{prop}\label{prop2} \vspace{0.02in}
Given $E_{j} \in \mathbb{R}^{n \times N}$ and $c_{j} \in \mathbb{R}^{N}$, a global minimizer of the dictionary atom update problem \eqref{eqop6} is
\begin{equation} \label{tru1ch4g}
\hat{d}_{j} =  \left\{\begin{matrix}
\frac{E_{j}c_{j}}{\left \| E_{j}c_{j} \right \|_{2}}, & \mathrm{if}\,\, c_{j}\neq 0 \\ 
v_{1}, & \mathrm{if}\,\, c_{j}= 0 
\end{matrix}\right.
\end{equation}
where $v_{1}$ is the first column of the $n \times n $ identity matrix. The solution is unique if and only if $c_{j}\neq 0$. 
\end{prop}

\hspace{0.1in} \textit{Proof:} 
First, for a vector $d_{j}$ that has unit $\ell_2$ norm, the following holds:
\begin{align} 
& \begin{Vmatrix}
E_{j} - d_{j}c_{j}^{T}
\end{Vmatrix}_{F}^{2}= \left \| E_{j} \right \|_{F}^{2} +  \left \| c_{j} \right \|_{2}^{2} - 2\, d_{j}^{T} E_{j} c_{j} 
\label{eqop12}
\end{align}
%By replacing the objective in problem \eqref{eqop6} with the right hand side in \eqref{eqop12}, it is clear that Problem \eqref{eqop6} is equivalent to finding the maximizer in
Upon substituting \eqref{eqop12} into \eqref{eqop6}, Problem \eqref{eqop6} simplifies to
\begin{equation} \label{eqop6b}
 \max_{d_{j}} \; d_{j}^{T} E_{j} c_{j}   \;\:\; \mathrm{s.t.}\; \: \left \| d_{j} \right \|_2 =1.
\end{equation}
By the Cauchy Schwarz inequality $d_{j}^{T} E_{j} c_{j} \leq \left \| E_{j} c_{j} \right \|_{2}$ for unit norm $d_{j}$. Thus, a solution to \eqref{eqop6b} that achieves the value $ \left \| E_{j} c_{j} \right \|_{2}$ for the objective is
\begin{equation} \label{tru1ch4g22}
\hat{d}_{j} =  \left\{\begin{matrix}
\frac{E_{j}c_{j}}{\left \| E_{j}c_{j} \right \|_{2}}, & \mathrm{if}\,\, E_{j}c_{j} \neq 0 \\ 
v_{1}, & \mathrm{if}\,\, E_{j}c_{j} = 0 
\end{matrix}\right.
\end{equation}
Obviously, any $d \in \mathbb{R}^{n}$ would be a minimizer (non-unique) in \eqref{eqop6b} when $E_{j}c_{j}=0$. In particular $\hat{d}_{j} =v_{1}$ works.

Next, we show that $E_{j}c_{j} = 0 $ in our algorithm if and only if $c_{j}=0$. This result together with \eqref{tru1ch4g22} immediately establishes the proposition. Since the $c_{j}$ used in the dictionary atom update step \eqref{eqop6} was obtained as a minimizer in the preceding sparse coding step \eqref{eqop5}, we have the following inequality for all $c \in \mathbb{R}^{N}$ with $\left \| c \right \|_{\infty} \leq L$, and $\tilde{d}_{j}$ denotes the $j$th atom in the preceding sparse coding step:
\begin{align} 
 & \hspace{-0.1in} \begin{Vmatrix}
E_{j} - \tilde{d}_{j}c_{j}^{T}
\end{Vmatrix}_{F}^{2} + \lambda^{2} \left \| c_{j} \right \|_{0}  \leq \begin{Vmatrix}
E_{j} - \tilde{d}_{j}c^{T}
\end{Vmatrix}_{F}^{2} + \lambda^{2} \left \| c \right \|_{0}
\label{eqop16b}
\end{align}
If $E_{j}c_{j}=0$, the left hand side above simplifies to $\left \| E_{j} \right \|_{F}^{2}$ $ + \left \| c_{j} \right \|_{2}^{2}$ $+ \lambda^{2} \left \| c_{j} \right \|_{0} $, which is clearly minimal when $c_{j}=0$. Thus, when $E_{j}c_{j}=0$, we must also have $c_{j}=0$.
$\;\;\; \blacksquare$

\begin{figure}
\begin{tabular}{p{8.3cm}}
\hline
SOUP-DIL Algorithm\\
\hline
 \textbf{Inputs\;:} \:\:\: Training data $ Y \in \mathbb{R}^{n \times N}$, weight $\lambda$, upper bound $L$, and number of iterations $K$.\\
 \textbf{Outputs\;:} \:\:\: Columns $ \left \{ d_{j}^{K} \right \}_{j=1}^{J}$ of the learned dictionary, and the learned sparse coefficients $\left \{c_{j}^{K} \right \}_{j=1}^{J}$. \\
\textbf{Initial Estimates:} $\left \{ d_{j}^{0}, c_{j}^{0} \right \}_{j=1}^{J}$.  (Often $c_{j}^{0} = 0$ $\forall$ $j$.)\\
\textbf{For \;$t$ = $1:$ $K$ Repeat}\\
\vspace{-0.09in}\hspace{0.01in} \textbf{For \;$j$ = $1:$ $J$ Repeat}
\begin{enumerate}
\item $C=\left [ c_{1}^{t},...,c_{j-1}^{t},c_{j}^{t-1},...,c_{J}^{t-1} \right ]$. \newline
$D=\left [ d_{1}^{t},...,d_{j-1}^{t},d_{j}^{t-1},...,d_{J}^{t-1} \right ]$. \vspace{0.04in}
\item \textbf{Sparse coding:} 
\begin{equation} \label{trree1}
b^{t} =   Y^{T} d_{j}^{t-1} - CD^{T}d_{j}^{t-1} + c_{j}^{t-1}
\end{equation}
\begin{equation} \label{tru1ch4g88bn}
c_{j}^{t} =  \min\left ( \begin{vmatrix}
H_{\lambda} \left ( b^{t}  \right )
\end{vmatrix}, L 1_{N} \right ) \, \odot \, \mathrm{sign}\left ( H_{\lambda} \left (  b^{t}  \right ) \right )
\end{equation}
%$c_{j}^{t} = H_{\lambda} \begin{pmatrix}
% \left ( E_{j}^{t} \right )^{T}d_{j}^{t-1} 
%\end{pmatrix}$
\item \textbf{Dictionary atom update:} 
\begin{equation} \label{rbb}
h^{t} = Yc_{j}^{t} - DC^{T}c_{j}^{t} + d_{j}^{t-1}\left ( c_{j}^{t-1} \right )^{T}c_{j}^{t}
\end{equation}
\begin{equation} \label{tru1ch4g88}
d_{j}^{t} =  \left\{\begin{matrix}
\frac{h^{t}}{\left \| h^{t} \right \|_{2}}, & \mathrm{if}\,\, c_{j}^{t}\neq 0 \\ 
v_{1}, & \mathrm{if}\,\, c_{j}^{t}= 0 
\end{matrix}\right.
\end{equation}
\end{enumerate}\\
\hspace{0.01in} \textbf{End} \\
\textbf{End} \\
\hline
\end{tabular}
\caption{The SOUP-DIL Algorithm (or SOUP-DILLO Algorithm, due to the use of the $\ell_{0}$ ``norm") for Problem (P1). Superscript of $t$ denotes the iterates in the algorithm. Although we perform the sparse coding step prior to the dictionary update step, one could also potentially switch this order. The vectors $b^{t}$ and $h^{t}$ above can be computed very efficiently via sparse operations.} \label{im5p}
%\vspace{-0.15in}
\end{figure}

While Propositions \ref{prop1} and \ref{prop2} provide the minimizers of \eqref{eqop5} and \eqref{eqop6} for the case of real-valued matrices/vectors in the problems, these solutions are trivially extended to the complex-valued case (that may be useful in applications such as magnetic resonance imaging \cite{bresai}) by 
%replacing the $(\cdot)^{T}$ in the propositions (and in P1) with $(\cdot)^{H}$, the 
using a Hermitian transpose. 
%operation.

Fig. \ref{im5p} shows the Sum of OUter Products DIctionary Learning (SOUP-DIL) Algorithm (or SOUP-DILLO Algorithm, due to the use of the $\ell_{0}$ ``norm") for Problem (P1). 
%We use "Algorithm 1" here to distinguish this method from the schemes in Part II. 
The algorithm assumes that an initial estimate $\left \{ d_{j}^{0}, c_{j}^{0} \right \}_{j=1}^{J}$ for the variables is provided. For example, the initial sparse coefficients could be set to zero, and the initial dictionary could be a known analytical dictionary such as the overcomplete DCT \cite{elad2}.
When $c_{j}^{t}= 0$, setting $d_{j}^{t} = v_{1}$ in \eqref{tru1ch4g88} in the algorithm could also be replaced with other (equivalent) settings such as $d_{j}^{t} = d_{j}^{t-1}$ or setting $d_{j}^{t}$ to a random unit norm vector. All of these settings have been observed to work well in practice. A random ordering of the atom/sparse coefficient updates in Fig. \ref{im5p} (i.e., random $j$ sequence) also helps in practice (in accelerating convergence) compared to cycling in the same order $1$ through $J$ every iteration. One could also alternate several times between the sparse coding and dictionary atom update steps for each $j$  in Fig.~\ref{im5p}. However, this would increase computation.

% The matrix-vector products above can be computed efficiently without explicitly computing the matrix $E_{j}^{t}$ each time or computing $E_{j}^{t}$ sequentially. This aspect is discussed in the text.

%$E_{j}^{t} = Y - \sum_{k<j} d_{k}^{t} \left ( c_{k}^{t} \right )^{T} - \sum_{k>j} d_{k}^{t-1} \left ( c_{k}^{t-1} \right )^{T}$

\subsection{Computational Cost} \label{sec3b} 

For each iteration $t$ in Fig. \ref{im5p}, our algorithm involves $J$ sparse code and dictionary atom updates. The sparse coding and atom update steps involve matrix-vector products for computing $b^{t}$ and $h^{t}$, respectively.
An alternative approach to the one in Fig. \ref{im5p} involves computing the matrix $E_{j}^{t} = Y - \sum_{k<j} d_{k}^{t} \left ( c_{k}^{t} \right )^{T} - \sum_{k>j} d_{k}^{t-1} \left ( c_{k}^{t-1} \right )^{T}$ (as in Propositions~\ref{prop1} and \ref{prop2}) directly at the beginning of each inner $j$ iteration.
This matrix can be updated sequentially and efficiently for each $j$ by adding and subtracting appropriate sparse rank-one matrices.
However, this alternative approach requires storing $E_{j}^{t} \in \mathbb{R}^{n \times N}$, which is often a large matrix for large $N$ and $n$. Instead, the procedure adopted by us in Fig. \ref{im5p} helps save memory usage.  We now discuss the cost of each sparse coding and atom update procedure in Fig. \ref{im5p}. %These products can be computed efficiently.

Consider the $t$th iteration and the $j$th inner iteration in  Fig.~\ref{im5p}, consisting of the update of the $j$th dictionary atom and its corresponding sparse coefficients. As in Fig. \ref{im5p}, let $D \in \mathbb{R}^{n \times J}$ be the dictionary whose columns are the current estimates of the atoms (at the start of the $j$th inner iteration), and let $C \in \mathbb{R}^{N \times J}$ be the corresponding sparse coefficients matrix. (The index $t$ on $D$ and $C$ is dropped to keep the notation simple.) Assume that the matrix $C$ has $\alpha N n$ non-zeros, with $\alpha \ll 1$ typically. This translates to an average of $\alpha n$ non-zeros per row of $C$ or $\alpha N n / J$ non-zeros per column of $C$. We refer to $\alpha$ as the sparsity factor of $C$.

The sparse coding step involves computing the right hand side of \eqref{trree1}.
%\begin{equation} \label{eqop21}
%\left ( E_{j}^{t} \right )^{T} d_{j}^{t-1} = Y^{T} d_{j}^{t-1} - CD^{T}d_{j}^{t-1} + c_{j}^{t-1}
%\end{equation}
While computing $Y^{T} d_{j}^{t-1} $ in \eqref{trree1} requires $N n$ multiply-add operations, computing $CD^{T}d_{j}^{t-1}$ using matrix-vector products requires $J n + \alpha N n$ multiply-add operations. Computing the difference of these vectors and summing that result with the sparse $c_{j}^{t-1}$ requires less than $2N$ additions. The hard thresholding-type operation \eqref{tru1ch4g88bn} to obtain $c_{j}^{t}$ requires at most $2N$ comparisons.

Next, when $c_{j}^{t} \neq 0$, the dictionary atom update step requires computing the right hand side of \eqref{rbb}.
%\begin{equation} \label{eqop22}
%E_{j}^{t}c_{j}^{t} = Yc_{j}^{t} - DC^{T}c_{j}^{t} + d_{j}^{t-1}\left ( c_{j}^{t-1} \right )^{T}c_{j}^{t}
%\end{equation}
Since $c_{j}^{t}$ is sparse with say $r_{j}$ non-zeros, computing $Yc_{j}^{t}$ in \eqref{rbb} requires $n r_{j}$ multiply-add operations, and computing $DC^{T}c_{j}^{t}$ using matrix-vector products requires less than $J n + \alpha N n$ multiply-add operations. The cost of the remaining operations in \eqref{rbb} and for normalizing $h^{t}$ \eqref{tru1ch4g88} is negligible.

Thus, the net cost of the $J \geq n$ inner iterations in iteration $t$ in Fig. \ref{im5p} is dominated (for $N \gg J, n$) by $N J n + 2 \alpha_{m} N J n + \beta  N n^{2}$, where $\alpha_{m}$ is the maximum sparsity factor of the estimated $C$'s during the inner iterations, and $\beta$ is the sparsity factor of the estimated $C$ at the end of iteration $t$. Thus, the cost per iteration of the block coordinate descent SOUP-DIL Algorithm is about $(1 + \alpha') N J n$, with $\alpha' \ll 1$ typically. On the other hand, the proximal alternating algorithm proposed very recently\footnote{This method also involves more parameters than our scheme.} by Bao et al. \cite{bao2, bao1} has a per-iteration computational cost of at least $2 N J n + 6 \alpha N J n + 4 \alpha N n^{2}$. This is clearly more computation than SOUP-DIL. 

Assuming $J \propto n$, the cost per iteration of the SOUP-DIL Algorithm scales as $O(N n^{2})$.
This is lower than the per-iteration cost of learning an $n \times J$ synthesis dictionary $D$ using K-SVD \cite{elad}, which scales\footnote{When $s \propto n$ and $J \propto n$, the per-iteration computational cost of the efficient implementation of K-SVD \cite{ge35b} also scales similarly as $O(Nn^{3})$.} (assuming that the synthesis sparsity level $s \propto n$ and $J \propto n$ in K-SVD) as $O(Nn^{3})$. 

As illustrated in Section \ref{sec5b}, our algorithm converges in few iterations in practice. Therefore, the per-iteration computational advantages also translate to net computational advantages in practice. The low computational cost of our approach could be particularly useful for big data applications, or higher dimensional (3D or 4D) applications.

\section{Convergence Analysis} \label{sec4}

This section presents a convergence analysis for the proposed SOUP-DIL Algorithm for Problem (P1). Problem (P1) is highly non-convex due to the $\ell_{0}$ penalty for sparsity, the unit $\ell_{2}$ norm constraints on atoms of $D$, and the term $\begin{Vmatrix}
Y- \sum_{j=1}^{J} d_{j}c_{j}^{T}
\end{Vmatrix}_{F}^{2}$ that is a non-convex function involving the products of multiple unknown vectors. The proposed algorithm is an exact block coordinate descent procedure for Problem (P1).  However, due to the high degree of non-convexity, standard results on convergence of block coordinate descent methods (e.g., \cite{tseng6}) do not apply here. More recent works \cite{xu222} on the convergence of block coordinate descent schemes use assumptions (such as multi-convexity) that do not hold in our setting.
Here, we discuss the convergence of our algorithm to the critical points in the problem. In the following, we first present some definitions and notations, before stating the main convergence results.

\subsection{Definitions and Notations} \label{sec4a}

\newtheorem{definition}{Definition}

First, we review the Fr\'{e}chet sub-differential of a function \cite{vari1, vari2}. The norm and inner product notation in Definition~\ref{def1} correspond to the euclidean $\ell_{2}$ settings.

\begin{definition} \label{def2}
For a function $g: \mathbb{R}^{p} \mapsto (-\infty, + \infty]$, its domain is defined as $\mathrm{dom} g = \left \{ x \in \mathbb{R}^{p} : g(x) < + \infty \right \}$. Function $g$ is proper if $\mathrm{dom} g$ is nonempty.
\end{definition}

\begin{definition} \label{def1}
Let $g: \mathbb{R}^{p} \mapsto (-\infty, + \infty]$ be a proper function and let $x \in \mathrm{dom} g$. The Fr\'{e}chet sub-differential of the function $g$ at $x$ is the following set denoted as $\hat{\partial }g(x) $:
\begin{equation*}
 \begin{Bmatrix}
h \in \mathbb{R}^{p} : \underset{b \to x, b \neq x}{\lim \inf} \frac{1}{\left \| b-x \right \|}\left ( g(b) - g(x) - \left \langle b-x, h \right \rangle \right ) \geq 0
\end{Bmatrix}
\end{equation*}
If $x \notin \mathrm{dom} g$, then $\hat{\partial }g(x) \triangleq \emptyset$, the empty set.
The sub-differential of $g$ at $x$ is the set $\partial g(x)$ defined as
\begin{equation*}
  \begin{Bmatrix}
\tilde{h}  \in \mathbb{R}^{p} : \exists x_{k} \to x, g(x_{k}) \to g(x), h_{k} \in \hat{\partial }g(x_{k}) \to \tilde{h} 
\end{Bmatrix}.
\end{equation*}
\end{definition}
A necessary condition for $x \in \mathbb{R}^{p}$ to be a minimizer of the function $g$ is that $x$ is a \emph{critical point} of $g$, i.e., $0 \in \partial g(x)$. 
Critical points are considered to be ``generalized stationary points" \cite{vari1}.
%If $g$ is a convex function, this condition is also sufficient.

We say that a sequence $ \left \{ z^{t} \right \}  \subset \mathbb{R}^{p}$ has an accumulation point $z$, if there is a subsequence that converges to $z$.

The constraints $\left \| d_{j} \right \|_2 =1$, $1 \leq j \leq J$, in (P1) can instead be added as penalties in the cost by using barrier functions $\chi (d_{j})$ (taking the value $+ \infty$ when the norm constraint is violated, and is zero otherwise). The constraints $\left \| c_{j} \right \|_{\infty} \leq L$, $1 \leq j \leq J$, can also be similarly replaced with barrier penalties $\psi(c_{j})$.
Then, we rewrite (P1) in unconstrained form with the following objective:
\begin{align} 
\nonumber & f(C, D) = f\left ( c_{1}, c_{2},..., c_{J}, d_{1}, d_{2},..., d_{J} \right ) =
\lambda^{2} \sum_{j=1}^{J} \left \| c_{j} \right \|_{0}\\
& \;\; + \begin{Vmatrix}
Y- \sum_{j=1}^{J} d_{j}c_{j}^{T}
\end{Vmatrix}_{F}^{2}  + \sum_{j=1}^{J} \chi (d_{j})  + \sum_{j=1}^{J} \psi(c_{j})  \label{eqop32}
\end{align}
For the SOUP-DIL Algorithm, the iterates computed in the $t$th outer iteration are denoted by the 2J-tuple $\left ( c_{1}^{t}, d_{1}^{t}, c_{2}^{t}, d_{2}^{t},..., c_{J}^{t}, d_{J}^{t} \right )$,
%$\left \{ c_{j}^{t}, d_{j}^{t} \right \}_{j=1}^{K}$, 
or alternatively by the pair of matrices $\left (  C^{t}, D^{t} \right )$.

% f\left ( \left \{ d_{k} \right \}, \left \{ c_{k} \right \} \right ) 

\subsection{Main Results} \label{sec4b}

Assume that the initial $(C^{0}, D^{0})$ satisfies the constraints in (P1). We then have the following simple monotonicity and consistency result.

\begin{theorem}\label{theorem1}\vspace{0.02in}
Let $\left \{ C^{t}, D^{t}\right \}$  denote the iterate sequence generated by the SOUP-DIL Algorithm with training data $Y \in \mathbb{R}^{n \times N}$ and initial $(C^{0}, D^{0})$. Then, the objective sequence  $\left \{ f^{t} \right \}$ with $f^{t} \triangleq f\left ( C^{t}, D^{t} \right )$ is monotone decreasing, and converges to a finite value, say $f^{*}$.
Moreover, the iterate sequence  $\left \{ C^{t}, D^{t} \right \}$ is bounded, and all its accumulation points are equivalent in the sense that they achieve the exact same value $f^{*}$ of the objective. 
\end{theorem}

\hspace{0.1in} \textit{Proof:} 
See Appendix \ref{app1}.
$\;\;\; \blacksquare$

% = f^{*}(C^{0}, D^{0})

Theorem \ref{theorem1} establishes that for each initial point $(C^{0}, D^{0})$, the bounded iterate sequence in the SOUP-DIL Algorithm is such that all its accumulation points achieve the same value $f^{*}$ of the objective. They are equivalent in that sense. The value of $f^{*}$ could vary with different initalizations.
We thus have the following corollary of Theorem \ref{theorem1} that holds because the distance between a bounded sequence and its (non-empty and compact) set of accumulation points  converges to zero.

\begin{cor}\label{corollaryt1a} \vspace{0.02in}
For each $(C^{0}, D^{0})$, the iterate sequence in the SOUP-DIL Algorithm converges to an equivalence class of accumulation points.
\end{cor}

The following Theorem \ref{theorem3} considers a very special case of SOUP dictionary learning, where the dictionary has a single atom. In this case, the SOUP learning Problem (P1) is the problem of obtaining a sparse rank-one approximation of the training matrix $Y$. In this case, Theorem \ref{theorem3} establishes that the iterates in the algorithm converge to the set of critical points (i.e., the distance between the iterates and the set converges to zero) of the objective $f$.

\begin{theorem}\label{theorem3} \vspace{0.02in}
Consider the SOUP-DIL Algorithm with $J=1$. Let $\left \{ c^{t}, d^{t} \right \}$ denote the bounded iterate sequence generated by the algorithm in this case with data $Y  \in \mathbb{R}^{n \times N}$ and initial $(c^{0}, d^{0})$. Then, every accumulation point of the iterate sequence is a critical point of the objective $f$, i.e., the iterates converge to the set of critical points of $f$.
\end{theorem}

\hspace{0.1in} \textit{Proof:} 
See Appendix \ref{app3}.
$\;\;\; \blacksquare$

For the general case ($J \neq 1$), we have the following result.
%In that case, the following Theorem establishes the critical point property of the accumulation points of the iterate sequence in Algorithm 1.

\begin{theorem}\label{theorem2} \vspace{0.02in}
Let $\left \{ C^{t}, D^{t} \right \}$ denote the bounded iterate sequence generated by the SOUP-DIL Algorithm with training data $Y  \in \mathbb{R}^{n \times N}$ and initial $(C^{0}, D^{0})$. Suppose each accumulation point $\left ( C, D\right )$ of the iterate sequence is such that the matrix $B$ with columns $b_{j} = E_{j}^{T}d_{j}$ and $E_{j} = Y - DC^{T} + d_{j}c_{j}^{T}$, has no entry with magnitude $\lambda$. Then every accumulation point of the iterate sequence is a critical point of the objective $f(C, D)$.
Moreover, the two sequences with terms $ \begin{Vmatrix}
D^{t} - D^{t-1}
\end{Vmatrix}_{F}$ and $ \begin{Vmatrix}
C^{t} - C^{t-1}
\end{Vmatrix}_{F}$ respectively,  both converge to zero.
\end{theorem}

\hspace{0.1in} \textit{Proof:} 
See Appendix \ref{app2}.
$\;\;\; \blacksquare$

%$ \begin{Vmatrix}D^{t}\left ( C^{t} \right )^{T} - D^{t-1}\left ( C^{t-1} \right )^{T} \end{Vmatrix}_{F}$

%$\left \{ a^{t} \right \}$ with $a^{t} \triangleq

Theorem \ref{theorem2} establishes that the iterates in SOUP-DIL converge to the set of critical points of $f(C, D)$.  For each initial $(C^{0}, D^{0})$, the iterate sequence in the algorithm converges (using Corollary \ref{corollaryt1a}) to an equivalence class of critical points of $f$.
Theorem \ref{theorem2} also establishes that $ \begin{Vmatrix}
D^{t} - D^{t-1}
\end{Vmatrix}_{F} \to 0$ and $ \begin{Vmatrix}
C^{t} - C^{t-1}
\end{Vmatrix}_{F} \to 0$, thereby implying that the sparse approximation to the training data $Z^{t}=D^{t}\left ( C^{t} \right )^{T}$ is such that $\begin{Vmatrix}
Z^{t} - Z^{t-1} 
\end{Vmatrix}_{F} \to 0$.  These are necessary but not sufficient conditions for the convergence of the entire sequences $\left \{ D^{t} \right \}$, $\left \{ C^{t} \right \}$, and $\left \{ Z^{t} \right \}$.
The assumption on the entries of the matrix $B$ in Theorem~\ref{theorem2} is equivalent to assuming that for every $1\leq j \leq J$, there is a unique minimizer of $f$ with respect to $c_{j}$ with all other variables fixed to their values in the accumulation point $(C, D)$.

Although Theorem \ref{theorem2} uses a uniqueness condition, the following conjecture postulates that provided the following Assumption 1 (that uses a probabilistic model for the data) holds, the uniqueness condition holds with probability $1$, i.e., the probability of a tie in assigning sparse codes is zero.

\textbf{Assumption 1.} The training signals $y_{i} \in \mathbb{R}^{n}$ for $1 \leq i \leq N$,  are  drawn independently from an absolutely continuous probability measure over the $n$-dimensional ball $S \triangleq \left \{ y \in \mathbb{R}^{n} : \left \| y \right \|_{2} \leq \beta_{0} \right \}$ for some $\beta_{0}>0$.

%Note that if Assumption 2 holds, then Assumption 1 also holds with probability 1. Alternatively, we can add a lower bound $\left \| y \right \|_{2} \geq c_1$ for some small $c_1>0$ in Assumption 3.

\newtheorem{conjecture}{Conjecture}

\begin{conjecture} \label{clstmetuneq} \vspace{0.02in}
Let Assumption 1 hold. Then, with probability 1, every accumulation point $\left ( C , D \right )$ of the iterate sequence in the SOUP-DIL Algorithm is such that for each $1\leq j \leq J$, the minimizer of $f(c_{1},...,c_{j-1},\tilde{c}_{j}, c_{j+1},...,c_{J}, d_{1},...,d_{J})$ with respect to $\tilde{c}_{j}$ is unique.
\end{conjecture}
\vspace{0.04in}

%satisfies
%\begin{equation*} %\label{aldso}
%(X, \Gamma) = \underset{\tilde{X}, \tilde{\Gamma}}{\arg\min} \; \,  g\left ( W, \tilde{X},\tilde{\Gamma} \right )
%\end{equation*}

If Conjecture \ref{clstmetuneq} holds, then every accumulation point of the iterate sequence in the SOUP-DIL Algorithm is immediately a critical point of $f(C, D)$ with probability $1$.

\section{Numerical Experiments}
\label{sec5}

\subsection{Framework} \label{sec5a}

This section presents numerical results illustrating the practical convergence behavior of the proposed algorithm, as well as its usefulness in sparse signal representation and image denoising. 
%In Part II, such experiments are conducted for algorithms that solve variants of Problem (P1). 
For demonstrating the convergence behavior and quality of sparse signal representations, we consider data formed using vectorized 2D patches of natural images. The SOUP-DIL Algorithm is used to learn sparse representations for such data, and we study the signal representation ability of the proposed algorithm at various values of $\lambda$ (i.e., various sparsities). We also present results for patch-based image denoising using Problem (P1). 
%We work with a patch based denoising scheme.

Our dictionary learning implementation was coded in Matlab version R2015a.
We compare the performance of dictionaries learned using our method to those learned using the classical K-SVD\footnote{The K-SVD method is a highly popular dictionary learning scheme that has been applied to many image processing applications including denoising \cite{elad2, elad3} and MR image reconstruction \cite{bresai}. 
Mairal et al. \cite{mai23} proposed a non-local method for image denoising that also exploits learned dictionaries and achieves denoising performance comparable to the well-known BM3D \cite{dbov} denoising method. Similar extensions to our proposed method for denoising and other applications are left for future work.} method  \cite{elad, elad2}.
For K-SVD learning and denoising \cite{elad, elad2}, we consider the original Matlab implementation of the methods available from Michael Elad's website \cite{el2}. A fast version\footnote{This version \cite{ge35} was observed to typically provide similar quality of results in our experiments as \cite{el2}.} of K-SVD \cite{ge35b} that also uses MEX/C implementations of sparse coding and some sub-steps of dictionary update, is publicly available \cite{ge35}. The Matlab implementation of our method is not currently optimized for efficiency.  Therefore, we compare the runtimes achieved by our unoptimized Matlab implementation to both the original Matlab and the efficient (partial MEX/C) implementations of K-SVD for a fair and instructive comparison. For K-SVD, we used the built-in parameter settings of the author's implementations, unless otherwise stated. All computations were performed with an Intel Xeon CPU X3230 at 2.66 GHz and 8 GB memory, employing a 64-bit Windows 7 operating system.

%Similar extensions to our proposed method to achieve state-of-the-art performance in denoising and other applications is left for future work.

For sparse representation of data $Y$, we use the \emph{normalized sparse representation error} (NSRE) $\left \| Y - DC^{T} \right \|_{F}/$ $\left \| Y \right \|_{F}$ to measure the performance of the learned dictonaries. For image denoising, similar to prior work, we measure the peak-signal-to-noise ratio (PSNR) computed between the true noiseless reference and the noisy or denoised images.   

\subsection{Convergence Experiment} \label{sec5b}

\begin{figure}[!t]
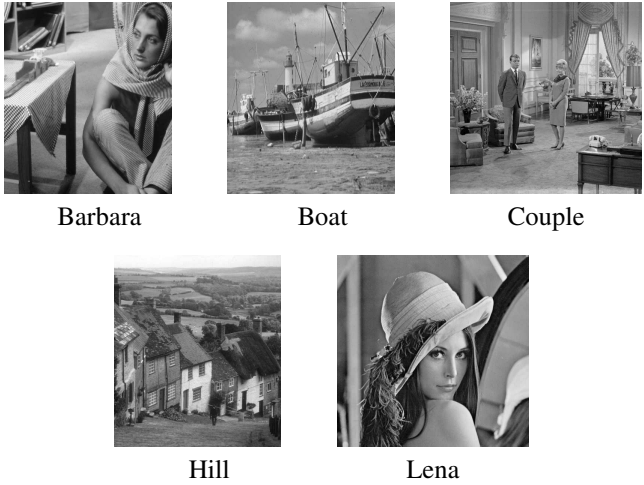

\begin{center}
\begin{tabular}{ccc}
\includegraphics[height=1.0in]{images/barbara}&
\includegraphics[height=1.0in]{images/boat}&
\includegraphics[height=1.0in]{images/couple}\\
Barbara & Boat & Couple \\
\vspace{-0.06in}
\end{tabular}
\begin{tabular}{cc}
\includegraphics[height=1.0in]{images/hill}&
\includegraphics[height=1.0in]{images/lena}\\
Hill & Lena \\
\end{tabular}
\caption{The $512 \times 512$ standard images used in our experiments. The images are Barbara, Boat, Couple, Hill, and Lena.}
\label{im6gch3}
\end{center}
%\vspace{-0.15in}
\end{figure}

To study the convergence behavior of the proposed SOUP-DIL Algorithm, we extracted $3 \times 10^{4}$ patches of size $8 \times 8$ from randomly chosen locations in the images Barbara, Boat, and Hill, shown in Fig. \ref{im6gch3}. The SOUP-DIL Algorithm for (P1) was then used to learn a $64 \times 256$ overcomplete dictionary, with $\lambda = 69$. The algorithm is initialized as mentioned in Section \ref{sec3a}. Specifically, the initial estimate for $C$ is an all-zero matrix, and the initial estimate for $D$ is the overcomplete DCT \cite{elad2, el2}.

\begin{figure}[!t]
\begin{center}
\begin{tabular}{cc}
\includegraphics[height=1.27in]{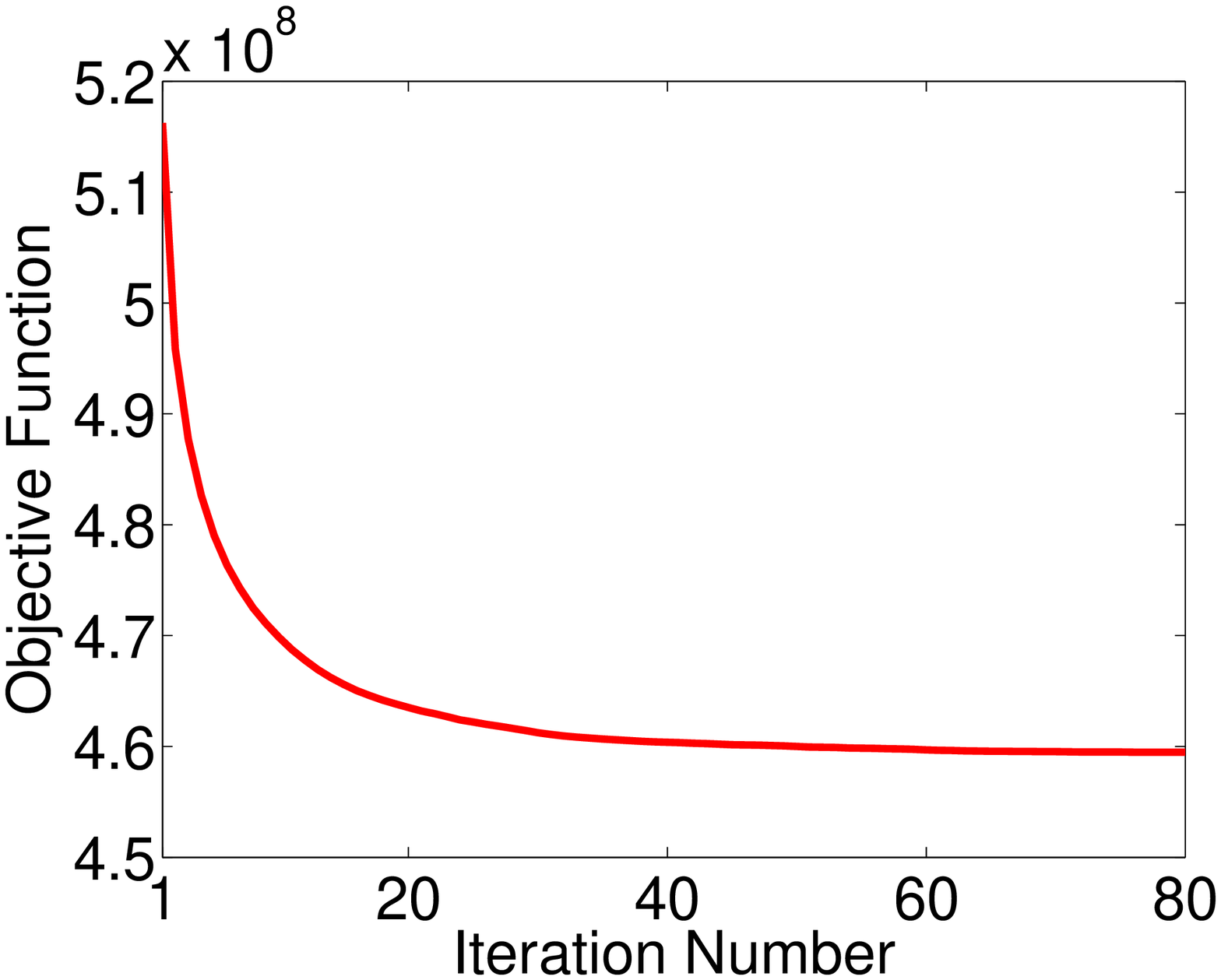}&
\includegraphics[height=1.2in]{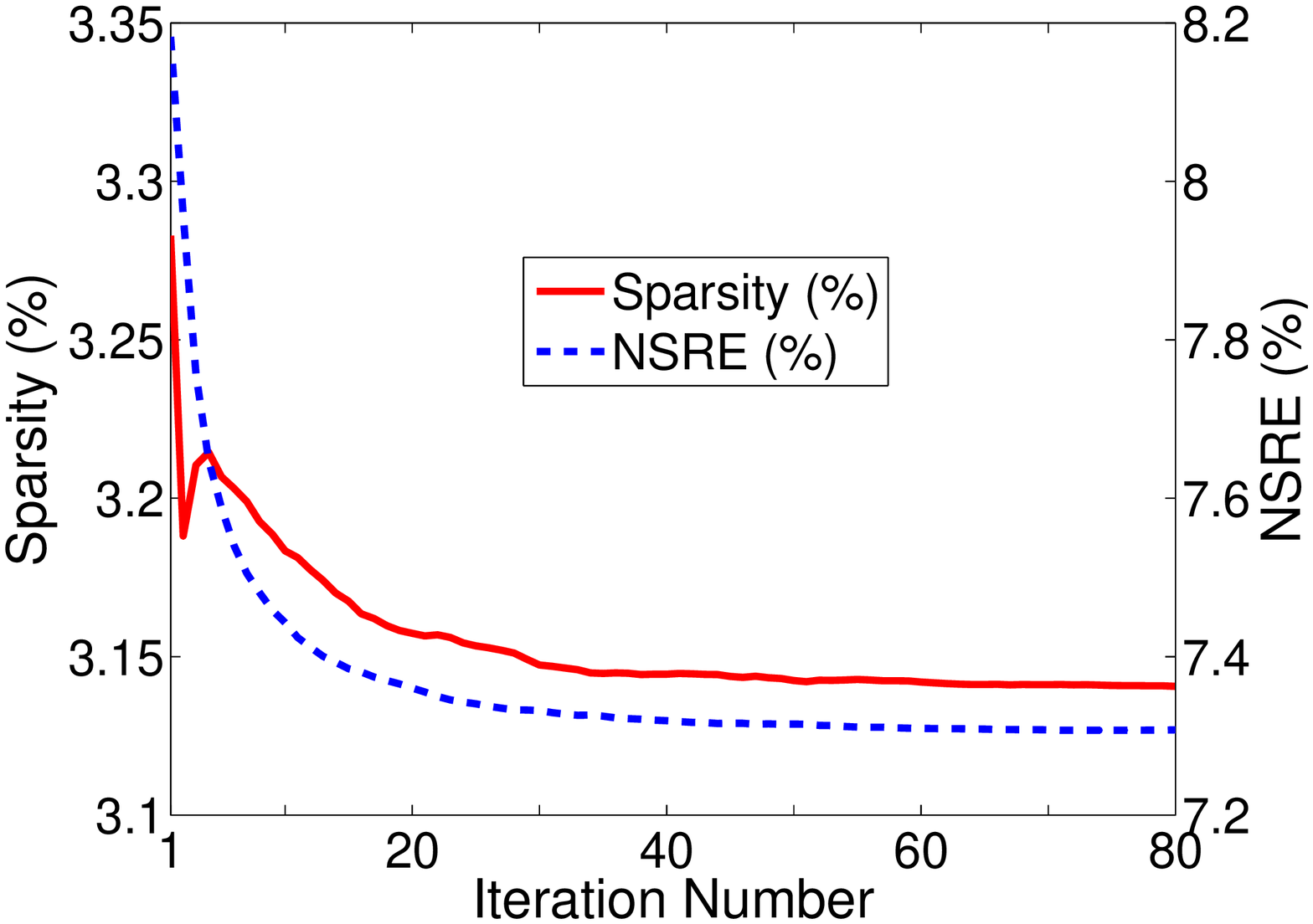}\\
(a) & (b) \\
\includegraphics[height=1.3in]{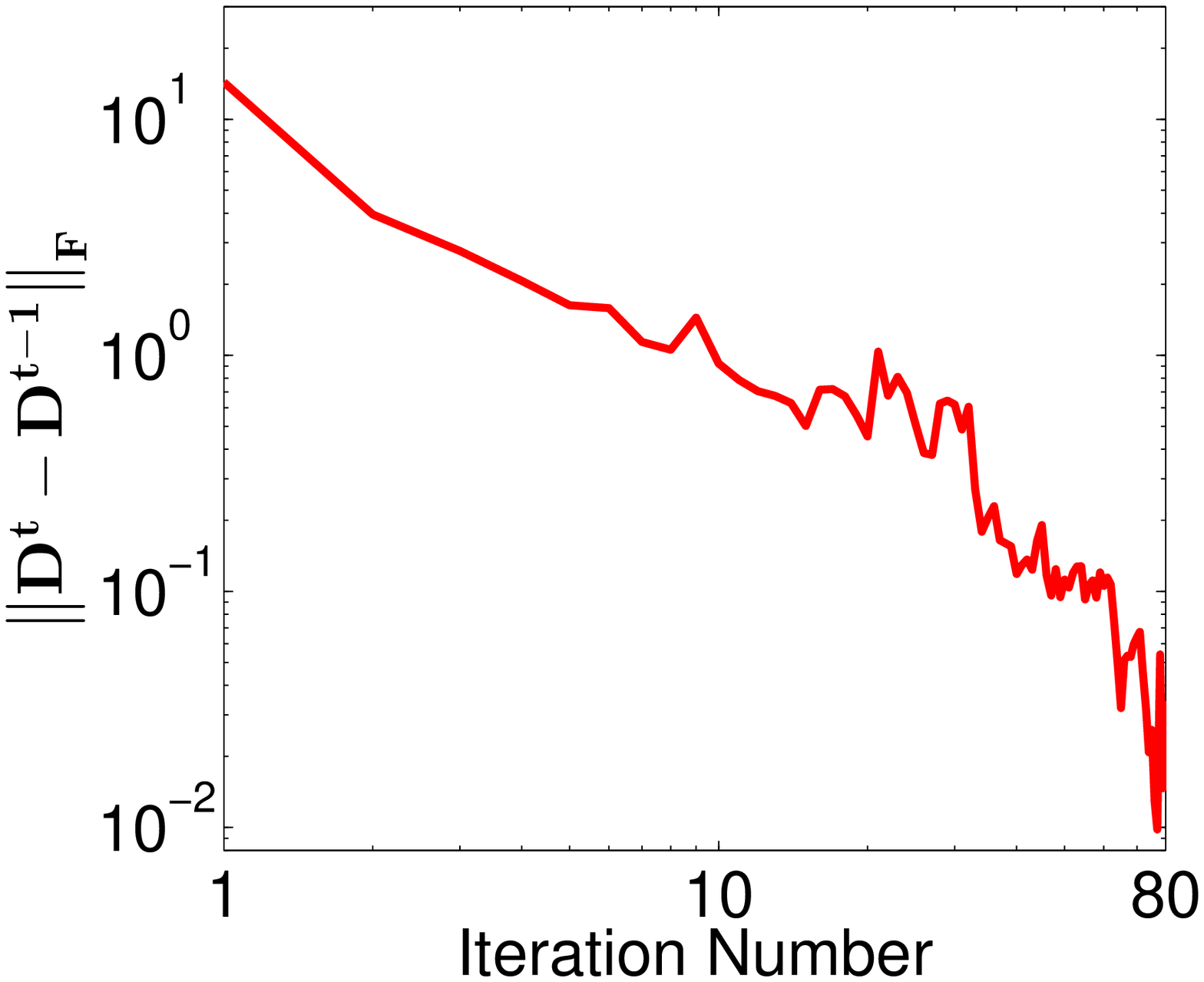}&
\includegraphics[height=1.38in]{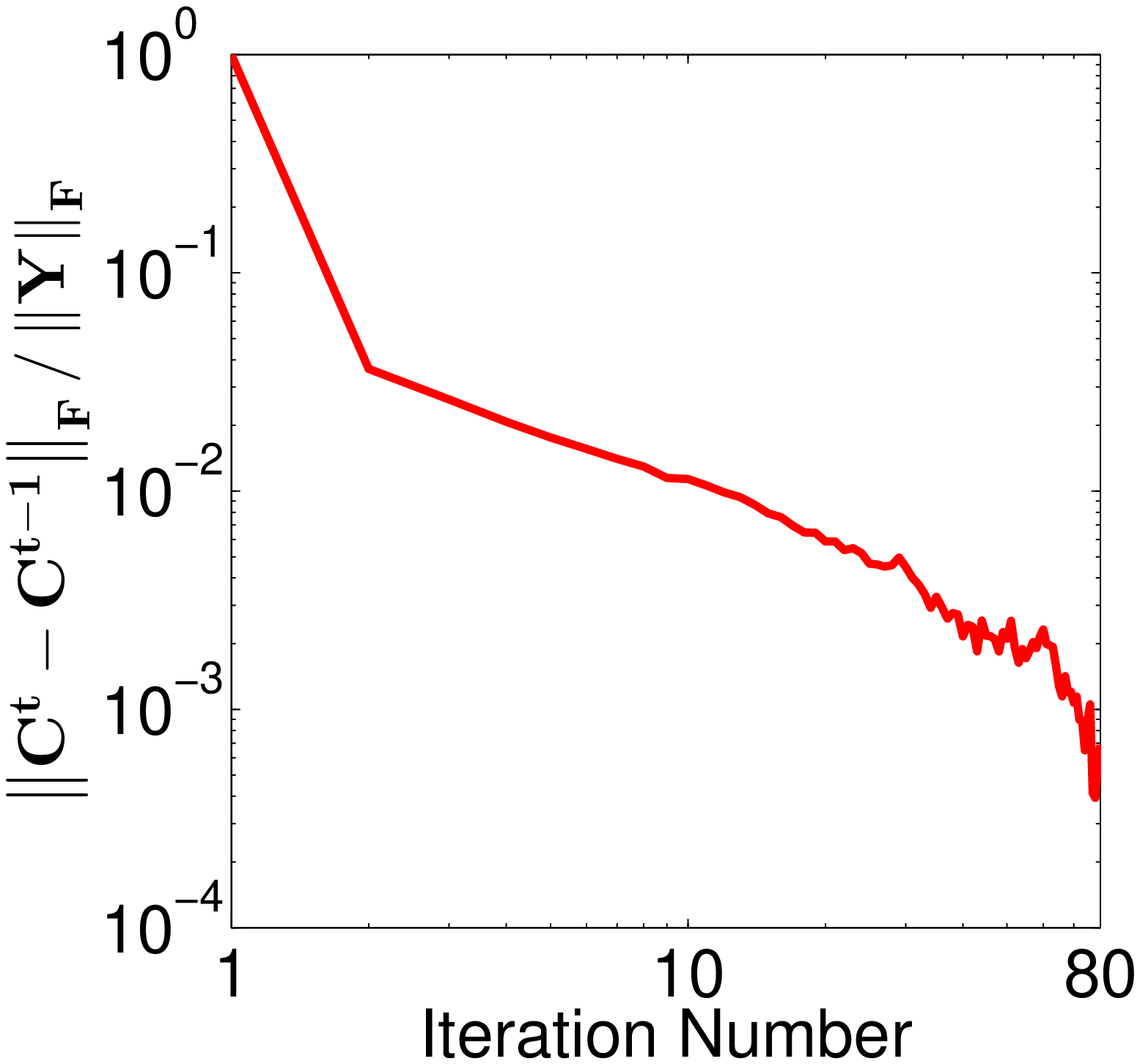}\\
(c) & (d) \\
\end{tabular}
\caption{Convergence behavior of the SOUP-DIL Algorithm: (a) Objective function; (b) Normalized sparse representation error (percentage) and sparsity factor of $C$ ($\sum_{j=1}^{J}\left ( \left \| c_{j} \right \|_{0} / nN \right )$ -- expressed as a percentage); (c) changes between successive $D$ iterates ($\left \| D^{t} - D^{t-1} \right \|_{F}$); and (d) normalized changes between successive $C$ iterates ($\left \| C^{t} - C^{t-1} \right \|_{F}/\left \| Y \right \|_{F}$).}
\label{im2}
\end{center}
%\vspace{-0.317in}
\end{figure}

 Fig. \ref{im2} illustrates the convergence behavior of SOUP-DIL. The objective in our method (Fig. \ref{im2}(a)) converged monotonically and quickly over the iterations.  Fig. \ref{im2}(b) shows the normalized sparse representation error and sparsity factor (for $C$), both expressed as percentages. Both these components of the objective converged quickly for the algorithm, and the NSRE improved by 1 dB beyond the first iteration, indicating the success of the SOUP-DIL approach in representing data using a small number of non-zero coefficients (sparsity factor of $3.14\%$ at convergence).

Importantly, both the quantities $\left \| D^{t} - D^{t-1} \right \|_{F}$ (Fig. \ref{im2}(c)) and $\left \| C^{t} - C^{t-1} \right \|_{F}$ (Fig. \ref{im2}(d)) converge towards $0$, as predicted by Theorem \ref{theorem2}.
This implies that $\begin{Vmatrix}
Z^{t} - Z^{t-1}
\end{Vmatrix}_{F} $ with $Z^{t}=D^{t} \left ( C^{t} \right )^{T}$ converges towards zero too.
The above results are indicative (are necessary but not sufficient conditions) of the convergence of the entire sequences $\left \{ D^{t} \right \}$,  $\left \{ C^{t} \right \}$, and $\left \{ Z^{t} \right \}$ for our algorithm in practice.
In contrast, Bao et al. \cite{bao1} showed that the distance between successive iterates may not converge to zero for popular algorithms such as K-SVD.

\subsection{Sparse Representation of Data} \label{sec5c}

\begin{figure}[!t]
\begin{center}
\begin{tabular}{cc}
\includegraphics[height=1.29in]{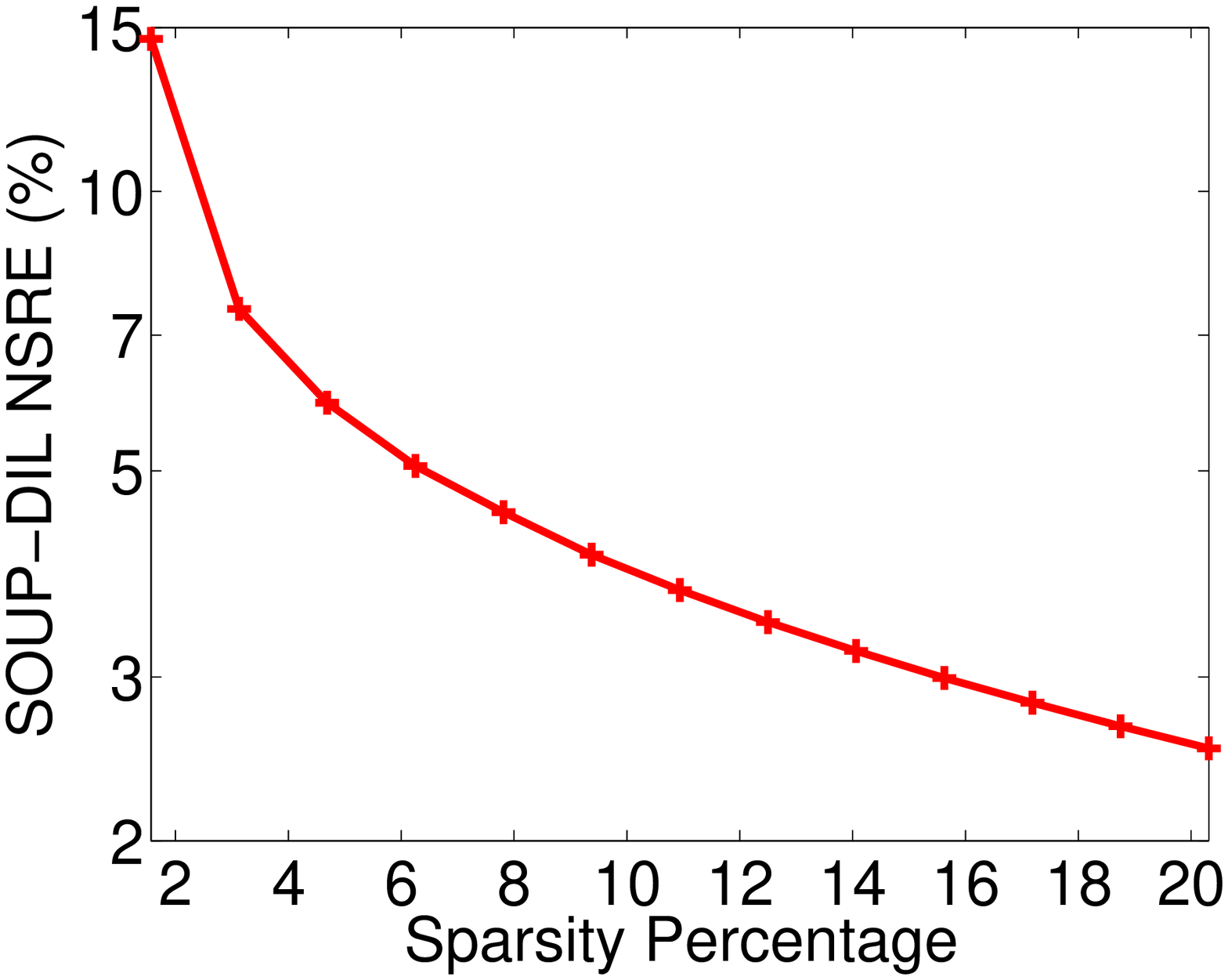}&
\includegraphics[height=1.27in]{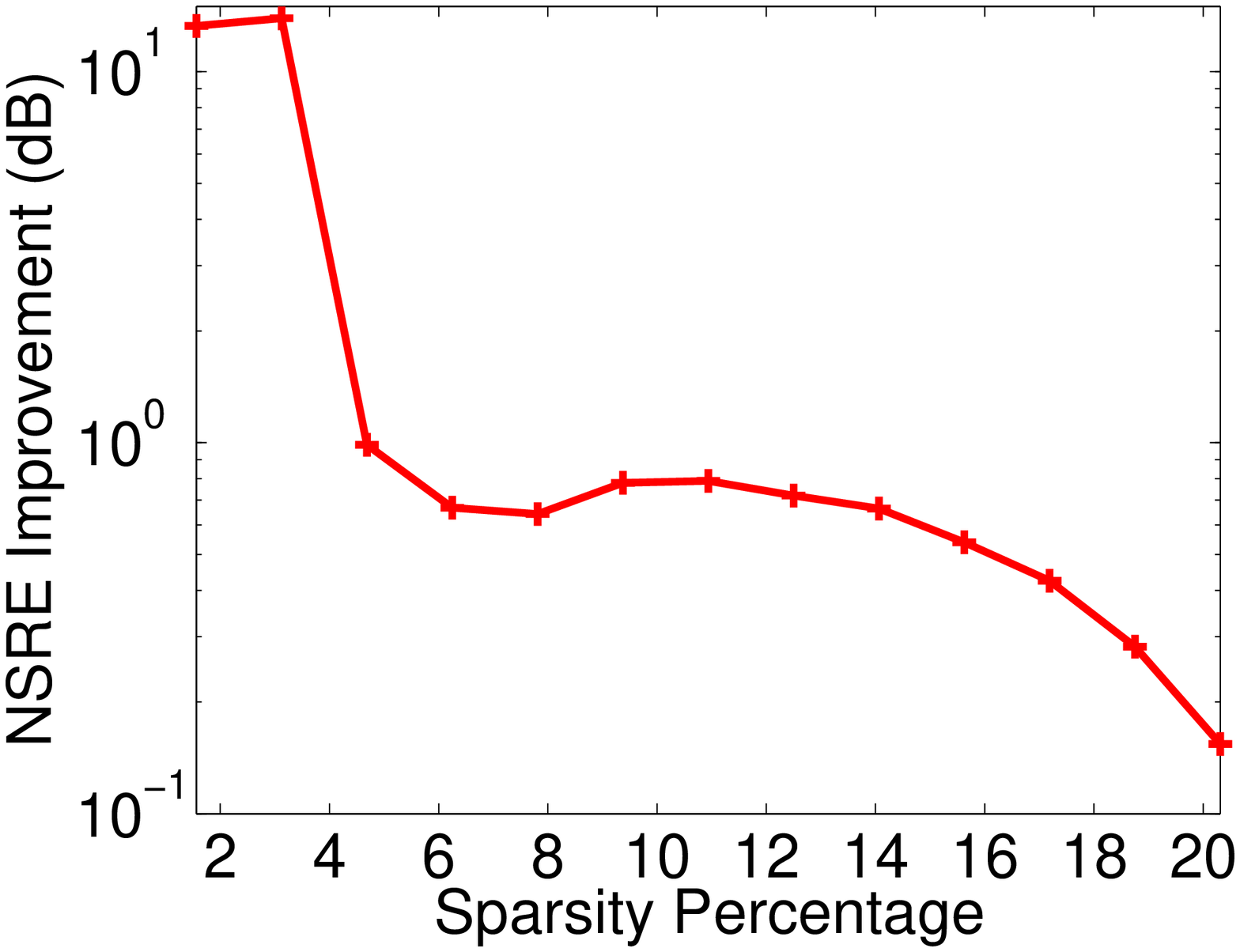}\\
(a) & (b) \\
\end{tabular}
\begin{tabular}{c}
\includegraphics[height=1.37in]{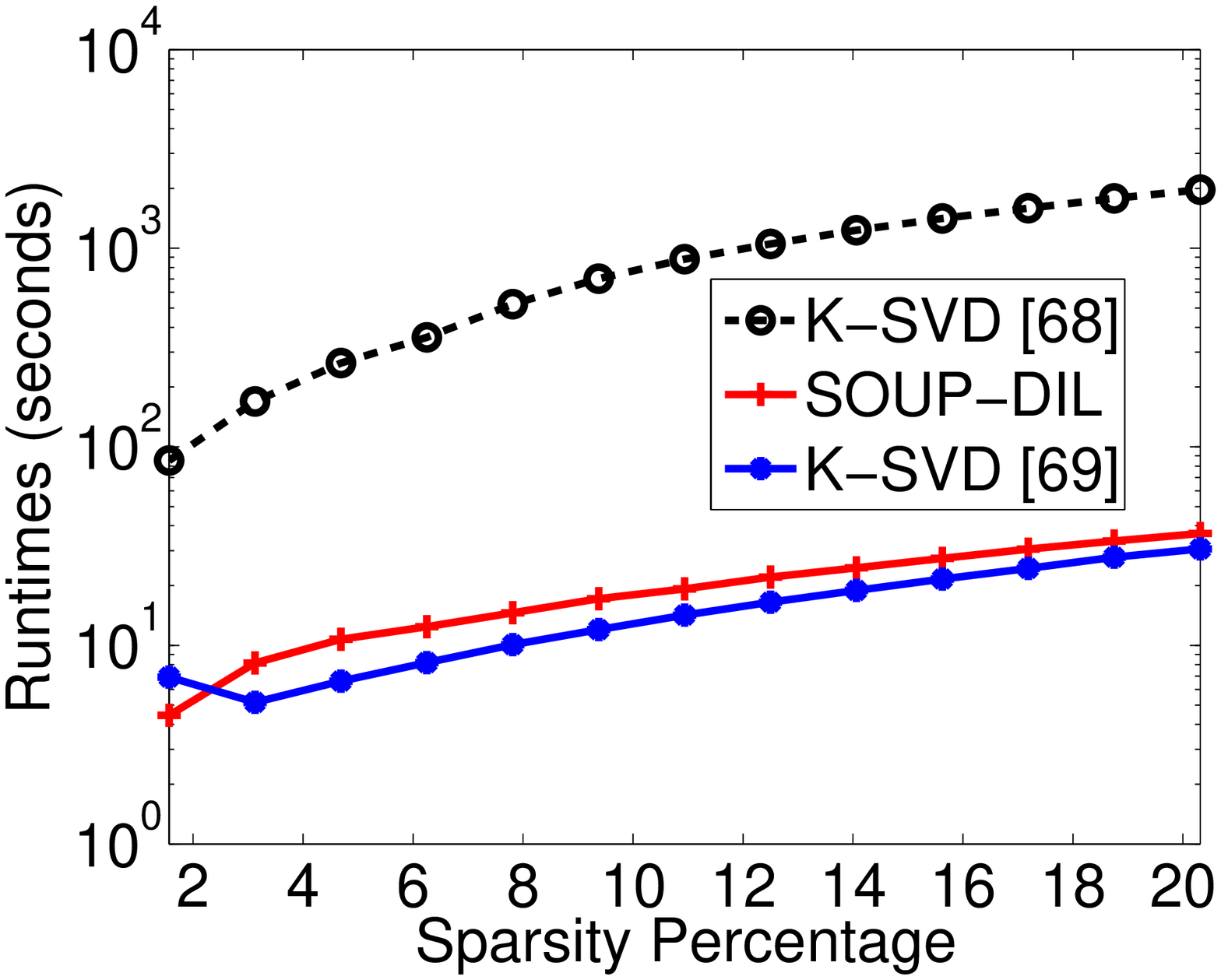}\\
(c) \\
\end{tabular}
\caption{Behavior of the SOUP-DIL Algorithm and K-SVD \cite{elad} at various sparsities (or sparsity factors): (a) NSRE (percentage) for SOUP-DIL; (b) Improvements in NSRE (in decibels) provided by SOUP-DIL over K-SVD; and (c) runtimes for SOUP-DIL, the matlab implementation of K-SVD \cite{elad, el2}, and the efficient (partial) MEX/C implementation of K-SVD \cite{ge35b, ge35}.}
\label{im3}
\end{center}
%\vspace{-0.317in}
\end{figure}

The second experiment worked with the same data as in Section \ref{sec5b} and learned dictionaries of size $64 \times 256$ for various choices of the parameter $\lambda$ in (P1) (i.e., corresponding to a variety of solution sparsity levels). We measure the quality of the learned data approximations $D
C^{T}$ using the NSRE metric. We compare the NSRE values achieved by our algorithm to those achieved by the K-SVD dictionary learning scheme \cite{elad} for the same data.
The K-SVD method was executed for several choices of sparsity (number of non-zeros) of the columns of  $C^{T}$. The $\lambda$ values for (P1) were chosen so as to achieve similar average column sparsity levels in $C^{T}$ as K-SVD.
Both the SOUP-DIL Algorithm and K-SVD were initialized with the same overcomplete DCT dictionary in our experiment and both methods ran for 10 iterations.

Fig. \ref{im3} shows the behavior of the SOUP-DIL and K-SVD algorithms for average column sparsity levels of $C^{T}$ ranging from about 2\% to 20\%.
As expected, the NSRE values for the SOUP-DIL Algorithm decreased monotonically (Fig. \ref{im3}(a)) when the average column sparsity level in $C^{T}$ increased (i.e., as $\lambda$ decreased). Importantly, the SOUP-DIL Algorithm provided better data representations (Fig. \ref{im3}(b)) than K-SVD at the various tested sparsity levels. Large improvements of about 14 dB and 1 dB are observed at low and mid sparsity levels, respectively.

Finally, Fig. \ref{im3}(c) compares the runtimes of our method to those of the unoptimized Matlab implementation of K-SVD \cite{el2} as well as the efficient (partial) MEX/C implementation \cite{ge35} of K-SVD demonstrating large speedups of 40-50 times for SOUP-DIL over the first  K-SVD implementation (at most sparsities), while the runtimes for the SOUP-DIL method are about the same as those of the second K-SVD implementation.
Since these results were obtained using only an unoptimized Matlab implementation of SOUP-DIL, we expect significant speed-ups for our scheme with code optimization or C/C++ implementations.

%While our method runs about 40-50 times faster than the 

\subsection{Image Denoising} \label{sec5d}

In image denoising, the goal is to recover an estimate of an image $x \in \mathbb{R}^{M}$ (2D image represented as a vector) from its corrupted measurements $y = x + h$, where $h$ is the noise. 
For example, the entries of $h$ may be i.i.d. Gaussian with zero mean and standard deviation $\sigma$.

To perform image denoising using (P1), we first extract all the overlapping patches (with maximum overlap) of the noisy image $y$, and construct the training set $Y \in \mathbb{R}^{n \times N}$ as a matrix whose columns are those noisy patches. We then use Problem (P1) to learn a dictionary and sparse codes for $Y$.
%and then obtain a denoised approximation to $Y$ as $\hat{D}\hat{C}^{T}$. 
To obtain the denoised image estimate, we then solve the following least squares problem, where $\hat{D}$ and $\hat{\alpha}_j$ denote the learned dictionary and patch sparse codes  obtained from the noisy patches, and $P_{j}$ is an operator that extracts a patch as a vector:
\begin{equation} \label{2efsw}
\min_{x} \sum_{j=1}^{N} \begin{Vmatrix}
P_{j}x - \hat{D} \hat{\alpha}_{j}
\end{Vmatrix}_{2}^{2}  +  \nu \left \| x-y \right \|_{2}^{2}
\end{equation}
The optimal $\hat{x}$ is obtained \cite{elad2} by summing together the denoised patch estimates $\hat{D} \hat{\alpha}_{j}$ at their respective 2D locations, and computing a weighted average between this result and the noisy image. 

\begin{table}[t]
\centering
\fontsize{8}{10pt}\selectfont
\begin{tabular}{|c|c|c|c|c|c|}
\hline
 Image & $\sigma$ & Noisy & O-DCT & K-SVD  & SOUP-DIL \\
\hline
\multirow{6}{*}{Couple}  &  5  &  34.16       &  37.25                 &  37.29              &   37.28                \\ \cline{2-6}
  &  10  & 28.11             & 33.40                             & 33.49                  &  33.50                     \\ \cline{2-6}
  &  20  & 22.11             & 29.71                             & 30.01                  &  29.99                    \\ \cline{2-6}
  &  25  &  20.17            & 28.53                             & 28.88                  &   28.92                    \\ \cline{2-6}
  &  30  &   18.58           & 27.53                             & 27.88                  &   27.97                     \\ \cline{2-6}
  &  100  &  8.13            & 22.59                             & 22.58                  &  22.71                      \\ 
\hline
\multirow{6}{*}{Barbara}  &  5  & 34.15      &   37.94               &  38.08              &  38.04                 \\ \cline{2-6}
  &  10  &  28.14            &  33.96                            & 34.43                  &  34.37                     \\ \cline{2-6}
  &  20  &  22.13            &  29.95                            & 30.83                  &  30.79                     \\ \cline{2-6}
  &  25  &  20.17            &  28.68                            & 29.63                  &  29.64                     \\ \cline{2-6}
  &  30  &   18.59           &  27.62                            & 28.54                  &  28.63                      \\ \cline{2-6}
  &  100  & 8.11             &  21.87                            & 21.87                  &  21.97                      \\ 
\hline
\multirow{6}{*}{Boat}  &  5  &  34.15          &    37.09               &  37.21              &   37.16                \\ \cline{2-6}
  &  10  &   28.13           & 33.43                             &  33.62                 &   33.60                    \\ \cline{2-6}
  &  20  &   22.10           & 29.92                             &  30.36                 &  30.37                     \\ \cline{2-6}
  &  25  &   20.17           & 28.79                             &  29.28                 &  29.30                     \\ \cline{2-6}
  &  30  &   18.60           & 27.93                             &  28.41                 &  28.43                      \\ \cline{2-6}
  &  100  &  8.13            & 22.79                             &  22.81                 &   22.96                     \\ 
\hline
\multirow{6}{*}{Hill}  &  5  & 34.15              &  37.02                 &  37.08              &  37.05                 \\ \cline{2-6}
  &  10  & 28.14             &  33.26                            & 33.45                  &   33.44                    \\ \cline{2-6}
  &  20  & 22.10             &  29.85                            & 30.17                  &    30.20                   \\ \cline{2-6}
  &  25  &  20.18            &  28.89                            & 29.23                  &    29.31                   \\ \cline{2-6}
  &  30  &  18.57            &  28.14                            & 28.43                  &    28.56                    \\ \cline{2-6}
  &  100  &  8.16            &  24.00                            & 23.98                  &    24.03                    \\ 
\hline
\multirow{6}{*}{Lena}  &  5  &   34.16         &  38.52                 &  38.62              &  38.55                 \\ \cline{2-6}
  &  10  & 28.12             &  35.30                            &  35.48                 &  35.47                     \\ \cline{2-6}
  &  20  & 22.11             &  32.02                            &  32.40                 &  32.40                     \\ \cline{2-6}
  &  25  & 20.18             &  30.89                            &  31.32                 &  31.32                     \\ \cline{2-6}
  &  30  & 18.59             &  29.98                            &  30.41                 &  30.46                      \\ \cline{2-6}
  &  100  & 8.14             &  24.45                           &   24.51                &    24.63                    \\ 
\hline
\multirow{6}{*}{Avg.}  &  5  &   34.16         &  37.56                 &  37.66              &  37.61                 \\ \cline{2-6}
  &  10  & 28.13             &  33.87                            &  34.09                 &  34.07                     \\ \cline{2-6}
  &  20  & 22.11             &  30.29                            &  30.75                 &  30.75                     \\ \cline{2-6}
  &  25  & 20.17             &  29.16                            &  29.67                 &  29.70                     \\ \cline{2-6}
  &  30  & 18.58             &  28.24                            &  28.74                 &  28.81                      \\ \cline{2-6}
  &  100  & 8.13             &  23.14                           &   23.15                &   23.26                    \\ 
\hline
\end{tabular}
\caption{PSNR values in decibels for denoising with overcomplete DCT (O-DCT), K-SVD \cite{elad2}, and SOUP-DIL. All dictionaries have size $64 \times 256$. The PSNR values of the noisy images (Noisy) are also shown. The last block of entries in the table corresponds to the average (over the images) PSNR values.}
\label{tabk1b}
%\vspace{-0.3in}
\end{table}

K-SVD based denoising \cite{elad2} involves a similar methodology as described above for (P1), but differs  in the dictionary learning procedure, where the $\ell_0$ ``norms" of the sparse codes are minimized so that a fitting constraint or error constraint of $\begin{Vmatrix}
P_{j}y-\hat{D}\hat{\alpha}_{j}
\end{Vmatrix}_{2}^{2} \leq \epsilon$ is met for representing the noisy patches. In particular, when the noise is i.i.d. Gaussian, $\epsilon = n C^{2} \sigma^{2}$ is used, with $C>1$ a constant. Such a constraint serves as a strong prior \cite{elad2, elad3}, and is a key reason for the denoising capability of K-SVD.
Hence, in our method based on (P1), we set $\lambda \propto \sigma$ (noise standard deviation) in (P1) during learning, and once the dictionary $\hat{D}$ is learned from noisy patches, we estimate the patch sparse codes $\hat{\alpha}_j$ using a single pass (over the noisy patches) of orthogonal matching pursuit (OMP) by employing an error constraint criterion like in K-SVD.
Although we only use information on the noise statistics in a sub-optimal way for (P1), we still show good denoising performance (vis-a-vis K-SVD) in the following with this approach. Note that for both our method and for K-SVD \cite{elad2, el2}, the means of the patches are removed prior to learning and added back afterwards to obtain the denoised estimates.

For the denoising experiments, we work with the images Couple, Barbara, Boat, Hill, and Lena in Fig. \ref{im6gch3}, and simulate i.i.d. Gaussian noise at six different noise levels ($\sigma = 5$, $10$, $20$, $25$, $30$, $100$) for each of the images. We then denoise these images using both the SOUP-DIL denoising method outlined above and K-SVD \cite{elad2, el2}.
SOUP-DIL denoising uses patches of size $8 \times 8$, a $64 \times 256$ overcomplete dictionary, $\lambda = 5 \sigma$, $\nu = 20/\sigma$ (in \eqref{2efsw}), an overcomplete DCT initial dictionary in learning, and 10 iterations of the learning algorithm. These parameter settings\footnote{The built-in parameter settings in the K-SVD denoising implementation \cite{elad2, el2} are very similar except that K-SVD uses a subset of the patches for training. We found that the denoising performance of K-SVD changes by at most few hundredths of a dB when all the patches are used in learning. However, this setting also leads to increased runtimes.} were found to work well for our method.

%The dictionary learning on the noisy patches (of size $8 \times 8$) for SOUP-DIL is executed with a $64 \times 256$ dictionary, $\lambda = 5 \sigma$, $\nu = 20/\sigma$ (in \eqref{2efsw}), an ove

\begin{figure}[!t]
\begin{center}
\begin{tabular}{cc}
\includegraphics[height=1.45in]{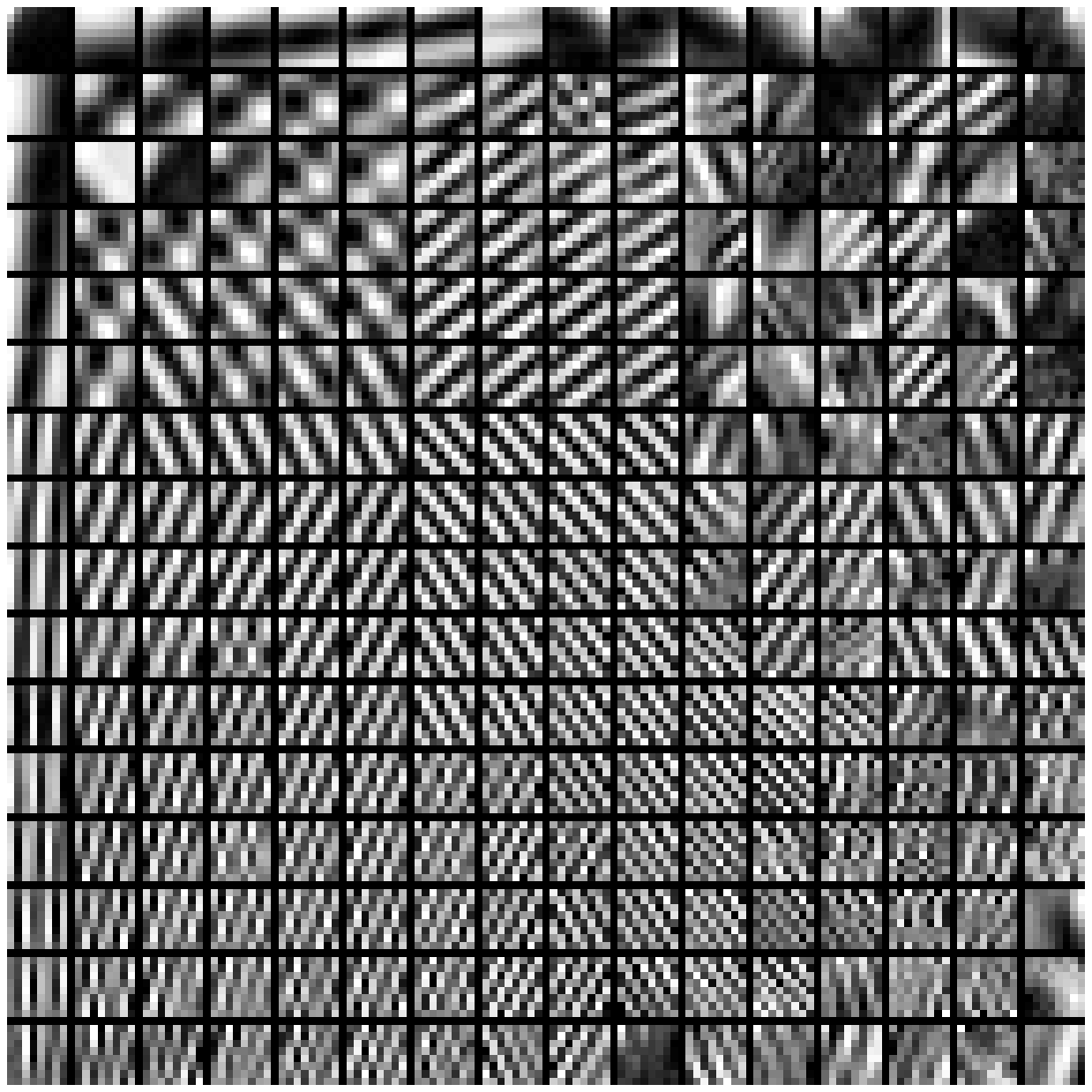}&
\includegraphics[height=1.45in]{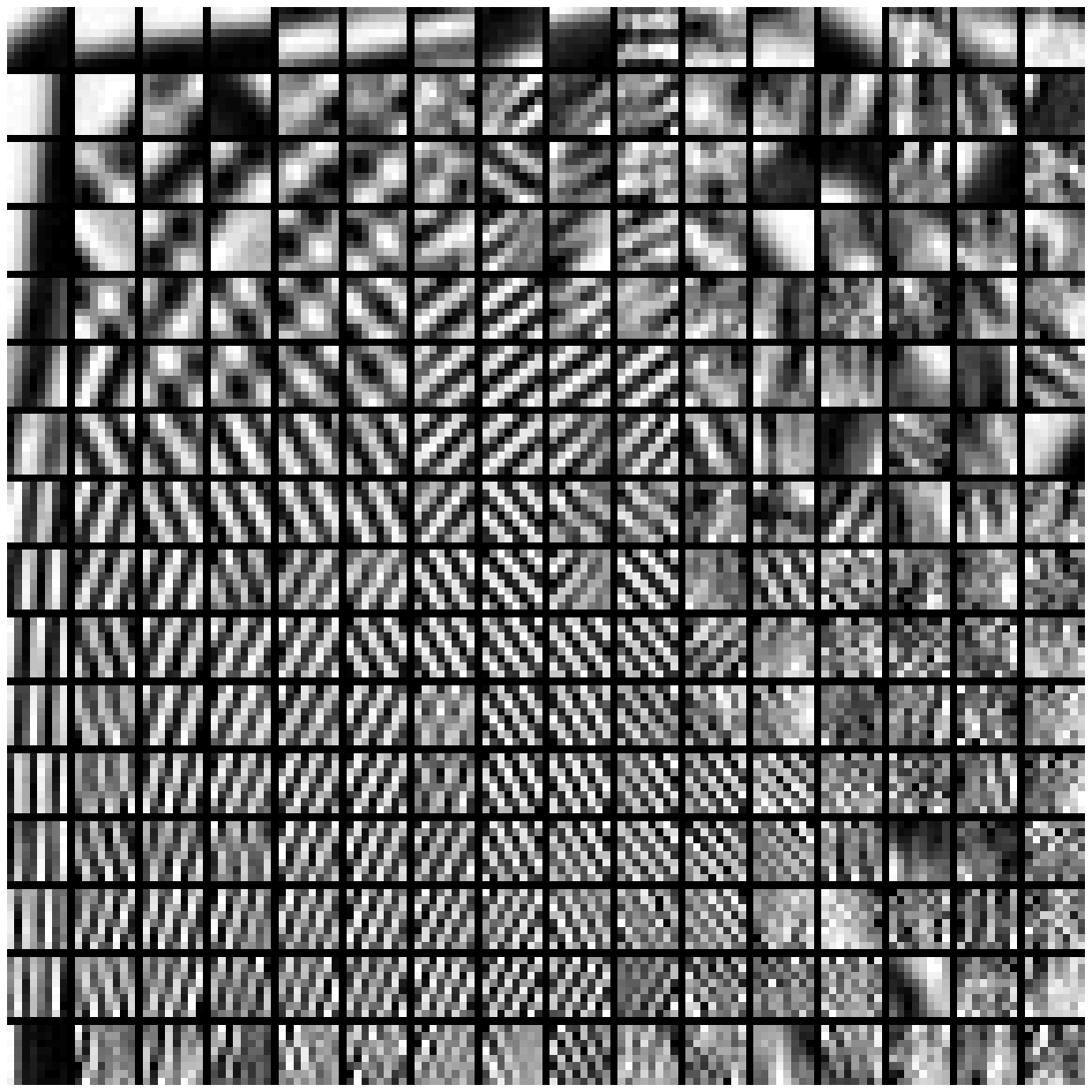}\\
(a) & (b) \\
\end{tabular}
\caption{Learned dictionaries for Barbara at $\sigma=20$: (a) SOUP dictionary; and (b) K-SVD dictionary.  The columns of the dictionaries are shown as patches.}
\label{im4}
\end{center}
%\vspace{-0.317in}
\end{figure}

Table \ref{tabk1b} lists the denoising PSNRs obtained by the SOUP-DIL denoising method, along with the PSNRs obtained by K-SVD. For comparison, we also list the denoising PSNRs obtained by employing the overcomplete DCT dictionary for denoising. In the latter case, the same strategy (and parameter settings \cite{el2}) as used by K-SVD based denoising is adopted but while skipping the learning process. 

The various methods denoise about the same in Table \ref{tabk1b} for a low noise level of $\sigma=5$. For $\sigma>5$, the SOUP-DIL method denoises about 0.4 dB better on the average than the overcomplete DCT. While the SOUP-DIL method and K-SVD denoise about the same at low and mid noise levels, the SOUP-DIL scheme performs about 0.1 dB better on average at higher noise levels such as $\sigma=30$ or $100$ in Table \ref{tabk1b}. Importantly, SOUP-DIL based denoising is highly efficient and the learning procedure has good convergence properties. We have observed similar effects as in Section \ref{sec5c} for the runtimes of SOUP-DIL denoising vis-a-vis the unoptimized Matlab \cite{el2} or the efficient (partial MEX/C) \cite{ge35} implementations of the K-SVD approach.

Fig. \ref{im4} shows the dictionaries learned from noisy patches using the SOUP-DIL method and K-SVD for the image Barbara at $\sigma=20$. Both dictionaries show frequency and textural features that are specific to the image Barbara. By learning such image-specific features, the SOUP-DIL method (and K-SVD) easily outperforms fixed dictionaries such as the overcomplete DCT in denoising.

%the denoised patch estimates are averaged together at their respective 2D locations and further

\section{Conclusions}
\label{sec6}  
This paper proposed a fast method for synthesis dictionary learning. The training data set is approximated by a sum of sparse rank-one matrices. The proposed formulation learns the left and right singular vectors of these rank-one matrices using an $\ell_0$ penalty to enforce sparsity. We adopted a block coordinate descent method to efficiently update the unknown sparse codes and dictionary atoms in the problem. In particular, the sparse coding step in our algorithm is performed cheaply by truncated  hard-thresholding, and the dictionary atom update involves a single sparse vector-matrix product that is computed efficiently. A convergence analysis was presented for the proposed block coordinate descent algorithm for a highly non-convex problem. The proposed approach had comparable or superior performance and significant speed-ups over the classical K-SVD method \cite{elad, elad2} in sparse signal representation and image denoising. The usefulness of our method in other applications such as blind compressed sensing \cite{sabsam} or other inverse problems in imaging merits further study. Extensions of the proposed method for online learning \cite{Mai} are also of potential interest.

%\vspace{-0.15in}

\appendices

\section{Proof of Theorem \ref{theorem1}} \label{app1}

First, we discuss the convergence of the objective sequence. At every iteration $t$ and inner iteration $j$ in Fig. \ref{im5p}, we solve the sparse coding and dictionary atom update subproblems exactly. Thus, we have the following two inequalities:
\begin{align} 
\nonumber& \hspace{-0.05in} f\left ( c_{1}^{t},..., c_{j-1}^{t}, c_{j}^{t}, c_{j+1}^{t-1},..., c_{J}^{t-1}, d_{1}^{t},...,d_{j-1}^{t}, d_{j}^{t-1},..., d_{J}^{t-1} \right ) \leq \\ 
\nonumber &   f\left ( c_{1}^{t},..., c_{j-1}^{t}, c_{j}^{t-1}, c_{j+1}^{t-1},..., c_{J}^{t-1}, d_{1}^{t},...,d_{j-1}^{t}, d_{j}^{t-1},..., d_{J}^{t-1} \right ) \\
\nonumber & f\left (c_{1}^{t},..., c_{j}^{t}, c_{j+1}^{t-1},..., c_{J}^{t-1}, d_{1}^{t},...,d_{j-1}^{t}, d_{j}^{t}, d_{j+1}^{t-1},..., d_{J}^{t-1} \right ) \leq \\
\nonumber & \;   f\left (c_{1}^{t},..., c_{j}^{t}, c_{j+1}^{t-1},..., c_{J}^{t-1}, d_{1}^{t},...,d_{j-1}^{t}, d_{j}^{t-1}, d_{j+1}^{t-1},..., d_{J}^{t-1} \right )
\end{align}
Combining these inequalities for the $t$th iteration and for each $1 \leq j \leq J$ yields $f(C^{t}, D^{t}) \leq$ $  f(C^{t-1}, D^{t-1})$. Thus, the sequence $\left \{ f(C^{t}, D^{t}) \right \}$ is monotone decreasing. Because it is lower bounded (by $0$), it converges to a finite value $f^{*} = f^{*}(C^{0}, D^{0})$ (that may depend on the initial conditions).

Next, the boundedness of the $\left \{ D^{t} \right \}$ and $\left \{ C^{t} \right \}$ sequences is obvious from the constraints in Problem (P1), so the accumulation points of the iterates form a non-empty and compact set. 

Finally, we show that each accumulation point achieves the same value $f^{*}$ of the objective.
Consider a subsequence $\left \{ C^{q_{t}}, D^{q_{t}} \right \}$ of the iterate sequence that converges to the accumulation point $\left ( C^{*}, D^{*} \right )$. Because $\left \| d_{j}^{q_t} \right \|_2 =1$ and $\left \| c_{j}^{q_t} \right \|_{\infty} \leq L$ for $1 \leq j \leq J$ and every $t$, therefore, due to the continuity of the norms, we also have $\left \| d_{j}^{*} \right \|_2 =1$ and $\left \| c_{j}^{*} \right \|_{\infty} \leq L$ for all $j$. Thus, the barrier functions in \eqref{eqop32} are inactive for the subsequence and its limit, i.e.,
\begin{equation} \label{ceyui2}
\chi(d_{j}^{*})=0, \, \psi(c_{j}^{*})=0 \,\; \forall \, j.
\end{equation}

The formula for $c_{j}^{t}$ in \eqref{tru1ch4g88bn} implies that it cannot have any non-zero entry of magnitude less than $\lambda$. Because $\left \{ c_{j}^{q_t} \right \}$ converges to $c_{j}^{*}$ entry-wise, therefore, for any entry of $c_{j}^{*}$ that is zero, the corresponding entry in $c_{j}^{q_t}= 0$ for large enough $t$.  Clearly, wherever $c_{j}^{*}$ is non-zero, the corresponding entry in $c_{j}^{q_t}$ stays non-zero for large enough $t$ values. Thus, 
\begin{equation} \label{vfev}
\lim_{t \to \infty} \begin{Vmatrix}
c_{j}^{q_t} 
\end{Vmatrix}_{0} = \begin{Vmatrix}
c_{j}^{*} 
\end{Vmatrix}_{0} \, \forall \, j
\end{equation} 
and the convergence in \eqref{vfev} happens in a finite number of iterations. We then easily have the following result:
\begin{align}
\nonumber & \lim_{t \to \infty} f(C^{q_t}, D^{q_t}) =  \lim_{t \to \infty} \begin{Vmatrix}
Y-  D^{q_t} \left ( C^{q_t} \right )^{T}
\end{Vmatrix}_{F}^{2}  \\ 
\nonumber &   +   \lambda^{2} \sum_{j=1}^{J} \lim_{t \to \infty} \left \| c_{j}^{q_t} \right \|_{0} = \lambda^{2} \sum_{j=1}^{J} \left \| c_{j}^{*} \right \|_{0}  + \begin{Vmatrix}
Y-  D^{*} \left ( C^{*} \right )^{T}
\end{Vmatrix}_{F}^{2}  %\label{csbldf00}
\end{align}
The right hand side above coincides with $ f(C^{*}, D^{*})$. Since the objective sequence converges to $f^{*}$, therefore, $f(C^{*}, D^{*})   = \lim_{t \to \infty} f(C^{q_t}, D^{q_t}) = f^{*}$. $\;\;\; \blacksquare$

\section{Proof of Theorem \ref{theorem3}} \label{app3}

First, consider a subsequence $\left \{ c^{q_t}, d^{q_t} \right \}$ of the iterate sequence that converges to the accumulation point $(c^{*}, d^{*})$. Because of the optimality of $d^{q_{t}}$ in the dictionary atom update step \eqref{eqop6} of the $q_{t}$th iteration, we have the following inequality for all $d \in \mathbb{R}^{n}$ with $\left \| d \right \|_{2}=1$:
\begin{equation} \label{cef}
\begin{Vmatrix}
Y - d^{q_t}\left ( c^{q_t} \right )^{T}
\end{Vmatrix}_{F}^{2} + \lambda^{2} \left \| c^{q_t} \right \|_{0} \leq \begin{Vmatrix}
Y - d \left ( c^{q_t} \right )^{T}
\end{Vmatrix}_{F}^{2} + \lambda^{2} \left \| c^{q_t} \right \|_{0}.
\end{equation}
Taking the limit $t \to \infty$ in \eqref{cef} and using \eqref{vfev} leads to the following result that holds for each feasible $d$:
\begin{equation*}
\begin{Vmatrix}
Y - d^{*}\left ( c^{*} \right )^{T}
\end{Vmatrix}_{F}^{2} + \lambda^{2} \left \| c^{*} \right \|_{0} \leq \begin{Vmatrix}
Y - d \left ( c^{*} \right )^{T}
\end{Vmatrix}_{F}^{2} + \lambda^{2} \left \| c^{*} \right \|_{0}.
\end{equation*}
The result implies that $d^{*}$ (which has unit $\ell_2$ norm) is in fact a global minimizer of $f(c^{*},d)$ with respect to $d$, i.e., $0 \in \partial  f_{d}\left ( c^{*}, d^{*}\right )$.

Next, consider a convergent subsequence $\left \{ c^{q_{n_t}+1} \right \}$ of the bounded sequence $\left \{ c^{q_{t}+1} \right \}$ that has a limit $c^{**}$. Because of the optimality of $ c^{q_{n_t}+1}$ in the sparse coding step \eqref{eqop5} of iteration $q_{n_t}+1$, we have the following inequality for all $c \in \mathbb{R}^{N}$ and for $f(c,d) =$ $ \begin{Vmatrix}
Y-dc^{T}
\end{Vmatrix}_{F}^{2} $ $+ \lambda^{2} \left \| c \right \|_{0} $ $+ \chi(d) + \psi(c)$: %with $\left \| c \right \|_{\infty} \leq L$:
\begin{equation}
f(c^{q_{n_{t}}+1},d^{q_{n_t}}) \leq f(c,d^{q_{n_t}})
\end{equation}
%\begin{align}
%\nonumber \begin{Vmatrix}
%Y - d^{q_{n_t}}\left ( c^{q_{n_t}+1} \right )^{T}
%\end{Vmatrix}_{F}^{2} + \lambda^{2} \left \| c^{q_{n_t}+1} \right \|_{0} \leq & \begin{Vmatrix}
%Y - d^{q_{n_t}}c^{T}
%\end{Vmatrix}_{F}^{2} \\
%\nonumber & + \lambda^{2} \left \| c \right \|_{0}.
%\end{align}
As in \eqref{cef}, taking the limit $t \to \infty$ above and using \eqref{vfev} and \eqref{ceyui2} yields the following result that holds for each $c$:
\begin{equation} \label{vfevwee}
f(c^{**},d^{*}) \leq f(c,d^{*})
\end{equation}
%\begin{equation*}
%\begin{Vmatrix}
%Y - d^{*}\left ( c^{**} \right )^{T}
%\end{Vmatrix}_{F}^{2} + \lambda^{2} \left \| c^{**} \right \|_{0} \leq \begin{Vmatrix}
%Y - d^{*}c^{T}
%\end{Vmatrix}_{F}^{2} + \lambda^{2} \left \| c \right \|_{0}.
%\end{equation*}
This implies that $c^{**}$ (that has $\left \| c^{**} \right \|_{\infty} \leq L$) is a global minimizer of $f(c, d^{*})$ with respect to $c$ achieving the objective value $f(c^{**}, d^{*})$. 

Since $f(c^{q_{n_t}+1},d^{q_{n_t}+1}) \leq$ $ f(c^{q_{n_t}+1},d^{q_{n_t}})$ $ \leq f(c^{q_{n_t}},c^{q_{n_t}})$, and  by Theorem \ref{theorem1}, the first and last terms in this inequality converge to $f^{*}$, we therefore have that $f(c^{**},d^{*})=f^{*}$ too. By Theorem \ref{theorem1}, $f(c^{*},d^{*})=f^{*}$ holds for the accumulation point $(c^{*}, d^{*})$ of the subsequence $\left \{ c^{q_t}, d^{q_t} \right \}$, implying $ f(c^{*},d^{*}) = f(c^{**},d^{*}) $.  Combining this with \eqref{vfevwee}, it is clear that $c^{*}$  is also a global minimizer of $f(c, d^{*})$ with respect to $c$, i.e.,  $0 \in \partial  f_{c}\left ( c^{*}, d^{*}\right )$.

Finally, (see Proposition 3 in \cite{Attouchaa}) the subdifferential $\partial  f $ at $\left ( c^{*}, d^{*} \right )$ satisfies $\partial  f\left ( c^{*}, d^{*}\right ) = 
\partial  f_{c}\left ( c^{*}, d^{*} \right )\times
\partial  f_{d}\left ( c^{*}, d^{*} \right )$.
%\begin{equation} \label{partlgob}
%\partial  f\left ( c^{*}, d^{*}\right ) = 
%\partial  f_{c}\left ( c^{*}, d^{*} \right )\times
%\partial  f_{d}\left ( c^{*}, d^{*} \right )
%\end{equation}
Since $0 \in \partial  f_{d}\left ( c^{*}, d^{*}\right )$ and $0 \in \partial  f_{c}\left ( c^{*}, d^{*}\right )$, we therefore have that $0 \in \partial  f\left ( c^{*}, d^{*} \right )$. Thus, when $J=1$ in the SOUP-DIL Algorithm, each accumulation point in the algorithm is a critical point of the objective $f$.
$\;\;\; \blacksquare$

\section{Proof of Theorem \ref{theorem2}} \label{app2}

\subsection{Critical Point Property} \label{app2a}

Consider a convergent subsequence $\left \{ C^{q_t}, D^{q_t} \right \}$ of the iterate sequence in the SOUP-DIL Algorithm that converges to $(C^{*}, D^{*})$. Let $\left \{ C^{q_{n_t}+1}, D^{q_{n_t} +1} \right \}$ be a convergent subsequence of the bounded $\left \{ C^{q_{t}+1}, D^{q_{t} +1} \right \}$, with limit $(C^{**}, D^{**})$.
For each iteration $t$ and inner iteration $j$ in the algorithm, define the matrix $E_{j}^{t} \triangleq Y - \sum_{k<j} d_{k}^{t} \left ( c_{k}^{t} \right )^{T}$ $ - \sum_{k>j} d_{k}^{t-1} \left ( c_{k}^{t-1} \right )^{T}$. For the accumulation point $(C^{*}, D^{*})$, define $E_{j}^{*} \triangleq Y - D^{*} \left ( C^{*} \right )^{T} $ $ + d_{j}^{*} \left ( c_{j}^{*} \right )^{T}$. In this proof, for simplicity, we denote the objective $f$ \eqref{eqop32} in the $j$th sparse coding step of iteration $t$ (Fig. \ref{im5p}) as 
\begin{align} 
\nonumber & f\left ( E_{j}^{t},c_{j}, d_{j}^{t-1} \right ) \triangleq \begin{Vmatrix}
E_{j}^{t}- d_{j}^{t-1}c_{j}^{T}
\end{Vmatrix}_{F}^{2} +
\lambda^{2} \sum_{k < j} \left \| c_{k}^{t} \right \|_{0} \\
&\;\;\; + \lambda^{2} \sum_{k > j} \left \| c_{k}^{t-1} \right \|_{0} + \lambda^{2} \left \| c_{j} \right \|_{0} + \psi(c_{j}) \label{eqopp20}
\end{align}
All but the $j$th atom and sparse vector $c_{j}$ are represented via $E_{j}^{t}$ on the left hand side in this notation. The objective that is minimized in the dictionary atom update step is similarly denoted as $f\left ( E_{j}^{t},c_{j}^{t}, d_{j} \right )$ with 
\begin{align} 
\nonumber & f\left ( E_{j}^{t},c_{j}^{t}, d_{j} \right )\triangleq \begin{Vmatrix}
E_{j}^{t}- d_{j}\left ( c_{j}^{t} \right )^{T}
\end{Vmatrix}_{F}^{2} +
\lambda^{2} \sum_{k \leq j} \left \| c_{k}^{t} \right \|_{0} \\
&\;\;\; + \lambda^{2} \sum_{k > j} \left \| c_{k}^{t-1} \right \|_{0} + \chi(d_{j}). \label{eqopp20bbb}
\end{align}
Finally, the functions $f\left ( E_{j}^{*},c_{j}, d_{j}^{*} \right )$ and $f\left ( E_{j}^{*},c_{j}^{*}, d_{j} \right )$ are defined in a similar way with respect to the accumulation point $(C^{*}, D^{*})$.

To establish the critical point property of $(C^{*}, D^{*})$, we first show the partial global optimality of each column of the matrices $C^{*}$ and $D^{*}$ for $f$. By partial global optimality, we mean that each column of $C^{*}$ or $D^{*}$ is a global minimizer of $f$, when all other variables are kept fixed to the values in $(C^{*}, D^{*})$.
First, for $j=1$ and iteration $q_{n_t} +1$, we have the following result for the sparse coding step for all $c_{1} \in \mathbb{R}^{N}$:
\begin{equation} \label{eqopp7}
f\left ( E_{1}^{q_{n_t}+1},c_{1}^{q_{n_t}+1}, d_{1}^{q_{n_t}} \right ) \leq f\left ( E_{1}^{q_{n_t}+1},c_{1}, d_{1}^{q_{n_t}} \right )
\end{equation}
Taking the limit $t \to \infty$ above and using \eqref{vfev} to obtain limits of $\ell_{0}$ terms in the cost \eqref{eqopp20}, and using \eqref{ceyui2}, and the fact that $E_{1}^{q_{n_t}+1} \to E_{1}^{*}$, we have
\begin{equation} \label{eqopp31}
f\left ( E_{1}^{*},c_{1}^{**}, d_{1}^{*} \right ) \leq f\left ( E_{1}^{*},c_{1}, d_{1}^{*} \right ) \, \forall \, c_{1} \in \mathbb{R}^{N}.
\end{equation} 
This means that $c_{1}^{**}$ is a minimizer of $f$ with all other variables fixed to their values in $(C^{*}, D^{*})$. Because of the (uniqueness) assumption in the theorem, we have
\begin{equation} \label{gobop1}
c_{1}^{**} = \underset{c_{1}}{\arg \min} \, f\left ( E_{1}^{*},c_{1}, d_{1}^{*} \right )
\end{equation}
Furthermore, because of the equivalence of accumulation points, $f\left ( E_{1}^{*},c_{1}^{**}, d_{1}^{*} \right )=$ $f\left ( E_{1}^{*},c_{1}^{*}, d_{1}^{*} \right )=f^{*}$ holds. This result together with \eqref{gobop1} implies that $c_{1}^{**}=c_{1}^{*}$ and thus
\begin{equation} \label{gobop1a}
c_{1}^{*} = \underset{c_{1}}{\arg \min} \, f\left ( E_{1}^{*},c_{1}, d_{1}^{*} \right )
\end{equation}
Therefore, $c_{1}^{*}$ is a partial global minimizer of $f$.

Next, for the first dictionary atom update step ($j=1$) in iteration $q_{n_t} +1$, we have the following for all $d_{1} \in \mathbb{R}^{n}$:
\begin{equation} \label{eqopp7b}
f\left ( E_{1}^{q_{n_t}+1},c_{1}^{q_{n_t}+1}, d_{1}^{q_{n_t}+1} \right ) \leq f\left ( E_{1}^{q_{n_t}+1},c_{1}^{q_{n_t}+1}, d_{1} \right )
\end{equation}
Just like in \eqref{eqopp7}, upon taking the limit $t \to \infty$ above and using $c_{1}^{**}=c_{1}^{*}$, we get 
\begin{equation} \label{eqopp31b}
f\left ( E_{1}^{*},c_{1}^{*}, d_{1}^{**} \right ) \leq f\left ( E_{1}^{*},c_{1}^{*}, d_{1} \right ) \, \forall \, d_{1} \in \mathbb{R}^{n}.
\end{equation} 
Thus, $d_{1}^{**}$ is a minimizer of the cost $ f\left ( E_{1}^{*},c_{1}^{*}, d_{1} \right )$ with respect to $d_{1}$. Because of the equivalence of accumulation points, we have $f\left ( E_{1}^{*},c_{1}^{*}, d_{1}^{**} \right )=$ $f\left ( E_{1}^{*},c_{1}^{*}, d_{1}^{*} \right )=f^{*}$. This implies that $d_{1}^{*}$ is also a partial global minimizer of $f$ in \eqref{eqopp31b} satisfying
\begin{equation} \label{gobop1b}
d_{1}^{*} \in \underset{d_{1}}{\arg \min} \, f\left ( E_{1}^{*},c_{1}^{*}, d_{1} \right ) 
\end{equation}
By Proposition \ref{prop2}, the minimizer of the dictionary atom update cost is unique as long as the corresponding sparse code (in \eqref{gobop1a}) is non-zero. Therefore, $d_{1}^{**}= d_{1}^{*}$ is the \emph{unique} minimizer in \eqref{gobop1b}, except when $c_{1}^{*}=0$. 

When  $c_{1}^{*}=0$, we use \eqref{vfev} to conclude that $c_{1}^{q_t}=0$ for all sufficiently large $t$ values. Since $c_{1}^{**}=c_{1}^{*}$, we must also have that $c_{1}^{q_{n_t}+1} = 0$ for all large enough $t$.
Therefore, for sufficiently large $t$, $d_{1}^{q_t}$ and $ d_{1}^{q_{n_t}+1} $ are the minimizers of dictionary atom update steps \eqref{eqop6}, wherein the corresponding sparse coefficients $c_{1}^{q_t}$ and $c_{1}^{q_{n_t}+1}$ are zero, implying that $d_{1}^{q_t} = d_{1}^{q_{n_t}+1} = v_{1} $ (Fig. \ref{im5p})  for all sufficiently large $t$. Thus, the limits satisfy $d_{1}^{*}= d_{1}^{**} = v_{1}$.
Therefore, $d_{1}^{**}= d_{1}^{*}$ holds, even when $c_{1}^{*} = 0$.
%\eqref{eqopp7b}
%in which case $d_{1}^{**}$ may differ from $d_{1}^{*}$. Nevertheless, in both these cases, the following result holds: 
Therefore, for $j=1$,
\begin{equation} \label{zebra1}
d_{1}^{**} =  d_{1}^{*}, \,\,\, c_{1}^{**} =  c_{1}^{*}.
\end{equation}
%d_{1}^{**}\left ( c_{1}^{**} \right )^{T}  = d_{1}^{**}\left ( c_{1}^{*} \right )^{T} = d_{1}^{*}\left ( c_{1}^{*} \right )^{T}.

Next, we repeat the above procedure by considering first the sparse coding step and then the dictionary update step for $j=2$ and iteration $q_{n_t} +1$.
For $j=2$, we consider the matrix
\begin{equation*}
E_{2}^{q_{n_t}+1} = Y - \sum_{k>2}d_{k}^{q_{n_t}}\left ( c_{k}^{q_{n_t}} \right )^{T}  - d_{1}^{q_{n_t}+1}\left ( c_{1}^{q_{n_t}+1} \right )^{T}
\end{equation*} 
It follows from \eqref{zebra1} that $E_{2}^{q_{n_t}+1}  \to E_{2}^{*}$ as $t \to \infty$. Then, by repeating the steps \eqref{eqopp7} - \eqref{zebra1}, we can easily show that $c_{2}^{*}$ and $d_{2}^{*}$ are each partial global minimizers of $f$ when all other variables are fixed to their values in $(C^{*}, D^{*})$. Moreover, $c_{2}^{**} =  c_{2}^{*}$ and $d_{2}^{**} =  d_{2}^{*}$. Similar such arguments can be made sequentially for all other values of $j$ until $j=J$.

Finally, the partial global optimality of each column of $C^{*}$ and $D^{*}$ for the objective $f$ implies that $0 \in \partial  f\left ( C^{*}, D^{*} \right )$, i.e., $(C^{*}, D^{*})$ is a critical point of the function $f$.
$\;\;\; \blacksquare$

\subsection{Convergence of the Difference between Successive Iterates} \label{app2b}
%Training Data Approximations

Consider the sequence $ \left \{ a^{t} \right \}$ whose elements are  $a^{t} \triangleq  \begin{Vmatrix}
D^{t} - D^{t-1} 
\end{Vmatrix}_{F} $. Clearly, this sequence is bounded. We will show that every convergent subsequence of this sequence converges to zero, thereby implying that zero is the only accumulation point, i.e., $ \left \{ a^{t} \right \}$ converges to 0. A similar argument establishes that $\begin{Vmatrix}
C^{t} - C^{t-1} 
\end{Vmatrix}_{F} \to 0$ as $t \to \infty$.

Consider a convergent subsequence $ \left \{ a^{q_{t}} \right \}$ of the sequence $ \left \{ a^{t} \right \}$.
The bounded sequence $\left \{ \left ( C^{q_{t}-1}, D^{q_{t}-1}, C^{q_{t}}, D^{q_{t}} \right ) \right \}$ (whose elements are formed by pairing successive elements of the iterate sequence)  must have a convergent subsequence $\left \{ \left ( C^{q_{n_t} - 1}, D^{q_{n_t} - 1}, C^{q_{n_t}}, D^{q_{n_t}} \right ) \right \}$ that converges to a point $(C^{*}, D^{*}, C^{**}, D^{**})$.
Based on the results in Appendix \ref{app2a}, we then have $d_{j}^{**} = d_{j}^{*} $ (and $c_{j}^{**} = c_{j}^{*} $) for each $1\leq j \leq J$, or
\begin{equation} \label{zebra2}
D^{**} = D^{*}.
\end{equation}
Thus, clearly $a^{q_{n_t}} \to 0$ as $t \to \infty$. 
Since, $ \left \{ a^{q_{n_t}} \right \}$ is a subsequence of the convergent $ \left \{ a^{q_{t}} \right \}$, we must have that $ \left \{ a^{q_{t}} \right \}$ converges to zero too. Thus, we have shown that zero is the limit of any arbitrary convergent subsequence of  $ \left \{ a^{t} \right \}$.
%Let $\left \{ C^{q_{n_t}+1}, D^{q_{n_t} +1} \right \}$ be a convergent subsequence of $\left \{ C^{q_{t}+1}, D^{q_{t} +1} \right \}$ with limit $(C^{**}, D^{**})$.
$\;\;\; \blacksquare$

\ifCLASSOPTIONcaptionsoff
  \newpage
\fi

% trigger a \newpage just before the given reference
% number - used to balance the columns on the last page
% adjust value as needed - may need to be readjusted if
% the document is modified later
%\IEEEtriggeratref{8}
% The "triggered" command can be changed if desired:
%\IEEEtriggercmd{\enlargethispage{-5in}}

% references section

% can use a bibliography generated by BibTeX as a .bbl file
% BibTeX documentation can be easily obtained at:
% http://www.ctan.org/tex-archive/biblio/bibtex/contrib/doc/
% The IEEEtran BibTeX style support page is at:
% http://www.michaelshell.org/tex/ieeetran/bibtex/
%\bibliographystyle{IEEEtran}
% argument is your BibTeX string definitions and bibliography database(s)
%\bibliography{IEEEabrv,../bib/paper}
%
% <OR> manually copy in the resultant .bbl file
% set second argument of \begin to the number of references
% (used to reserve space for the reference number labels box)

\bibliographystyle{./IEEEtran}
\bibliography{./IEEEabrv,./OPDL_v3}

%\begin{thebibliography}{1}

%\bibitem{IEEEhowto:kopka}
%H.~Kopka and P.~W. Daly, \emph{A Guide to \LaTeX}, 3rd~ed.\hskip 1em plus
%  0.5em minus 0.4em\relax Harlow, England: Addison-Wesley, 1999.

%\end{thebibliography}

% biography section
%
% If you have an EPS/PDF photo (graphicx package needed) extra braces are
% needed around the contents of the optional argument to biography to prevent
% the LaTeX parser from getting confused when it sees the complicated
% \includegraphics command within an optional argument. (You could create
% your own custom macro containing the \includegraphics command to make things
% simpler here.)
%\begin{biography}[{\includegraphics[width=1in,height=1.25in,clip,keepaspectratio]{mshell}}]{Michael Shell}
% or if you just want to reserve a space for a photo:

%\begin{IEEEbiography}{Michael Shell}
%Biography text here.
%\end{IEEEbiography}

% if you will not have a photo at all:
%\begin{IEEEbiographynophoto}{John Doe}
%Biography text here.
%\end{IEEEbiographynophoto}

% insert where needed to balance the two columns on the last page with
% biographies
%\newpage

%\begin{IEEEbiographynophoto}{Jane Doe}
%Biography text here.
%\end{IEEEbiographynophoto}

% You can push biographies down or up by placing
% a \vfill before or after them. The appropriate
% use of \vfill depends on what kind of text is
% on the last page and whether or not the columns
% are being equalized.

%\vfill

% Can be used to pull up biographies so that the bottom of the last one
% is flush with the other column.
%\enlargethispage{-5in}

% that's all folks
\end{document}